\documentclass{article}

\usepackage{microtype}
\usepackage{graphicx}
\usepackage{subfigure}
\usepackage{booktabs} %

\usepackage{hyperref}

\usepackage[accepted]{icml2024}

\usepackage{amsmath}
\usepackage{amssymb}
\usepackage{mathtools}
\usepackage{amsthm}
\usepackage{comment}

\usepackage[capitalize,noabbrev]{cleveref}

\theoremstyle{plain}

\theoremstyle{definition}

\theoremstyle{remark}

\usepackage[textsize=tiny]{todonotes}

\usepackage{ulem}
\usepackage{soul}
\usepackage{amsmath}
\usepackage{amssymb}
\usepackage{booktabs}
\usepackage{multirow}
\usepackage{bbm}
\usepackage{wrapfig}
\usepackage{makecell}
\usepackage{bbding}
\usepackage{pifont}
\usepackage{wasysym}
\usepackage{color, colortbl}
\usepackage{mathtools}

\usepackage{enumitem}

\usepackage{array}
\newcolumntype{x}[1]{>{\centering\arraybackslash\hspace{0pt}}p{#1}}

\newcommand{\hr}[1]{\textcolor{black}{#1}}

\newcommand{\tabincell}[2]{\begin{tabular}{@{}#1@{}}#2\end{tabular}} 

\icmltitlerunning{When Linear Attention Meets Autoregressive Decoding: Towards More Effective and Efficient Linearized LLMs}

\begin{document}

\twocolumn[
\icmltitle{When Linear Attention Meets Autoregressive Decoding:\\Towards More Effective and Efficient Linearized Large Language Models}

\icmlsetsymbol{equal}{*}

\begin{icmlauthorlist}
\icmlauthor{Haoran You}{GT}
\icmlauthor{Yichao Fu}{GT}
\icmlauthor{Zheng Wang}{GT}
\icmlauthor{Amir Yazdanbakhsh}{Google}
\icmlauthor{Yingyan (Celine) Lin}{GT}
\end{icmlauthorlist}

\icmlaffiliation{GT}{School of Computer Science, Georgia Institute of Technology, Atlanta, USA}
\icmlaffiliation{Google}{Google DeepMind, Mountain View, USA}

\icmlcorrespondingauthor{Haoran You}{haoran.you@gatech.edu}

\icmlkeywords{Machine Learning, ICML}

\vskip 0.3in
]

\printAffiliationsAndNotice{}  %

\begin{abstract}
Autoregressive Large Language Models (LLMs) have achieved impressive performance in language tasks but face two significant bottlenecks: (1) quadratic complexity in the attention module as the number of tokens increases, and (2) limited efficiency due to the sequential processing nature of autoregressive LLMs during generation. While linear attention and speculative decoding offer potential solutions, their applicability and synergistic potential for enhancing autoregressive LLMs remain uncertain. We conduct the first comprehensive study on the efficacy of existing linear attention methods for autoregressive LLMs, integrating them with speculative decoding. We introduce an augmentation technique for linear attention that ensures compatibility with speculative decoding, enabling more efficient training and serving of LLMs. Extensive experiments and ablation studies involving seven existing linear attention models and five encoder/decoder-based LLMs consistently validate the effectiveness of our augmented linearized LLMs. Notably, our approach achieves up to a 6.67 reduction in perplexity on the LLaMA model and up to a 2$\times$ speedup during generation compared to prior linear attention methods. Codes and models are available at \url{https://github.com/GATECH-EIC/Linearized-LLM}.
\end{abstract}

\section{Introduction}
\label{sec:intro}

LLMs have demonstrated exceptional capabilities in language understanding and generation tasks, sparking immense interest.
Autoregressive LLMs, like OpenAI's ChatGPT~\cite{chatgpt,gpt4}, Meta's LLaMA~\cite{touvron2023llama,touvron2023llama2}, and Google's Gemini~\cite{team2023gemini}, have achieved state-of-the-art (SOTA) performance in generation.
However, these models suffer from significant computational and memory demands, hindering their efficiency in both training and serving. These limitations stem from two key bottlenecks:
\textbf{\textit{Bottleneck 1}}: The attention module, a core component of LLMs, exhibits quadratic complexity relative to the input sequence length. This necessitates training LLMs with limited context sizes (e.g., 2048 tokens for LLaMA), restricting their ability to process lengthy documents or engage in extended conversations~\cite{chen2023longlora}.
\textbf{\textit{Bottleneck 2}}: The sequential nature of autoregressive decoding limits parallelism during generation, resulting in slow inference speeds, especially for long sequences~\cite{miao2023specinfer}.

Various techniques have been proposed to address these bottlenecks, including pruning~\cite{llm_pruning}, quantization~\cite{frantar2022optq,xiao2023smoothquant,qs-theory}, speculative decoding~\cite{miao2023specinfer,leviathan2023fast}, and linear attention~\cite{qin2023scaling,lu2021soft}.
Among these, linear attention tackles \textbf{\textit{Bottleneck 1}} by reducing the quadratic complexity of softmax attention from quadratic to linear.
Speculative decoding addresses \textbf{\textit{Bottleneck 2}} by employing
smaller draft models for speculative parallel generation, followed by verification using the full LLM~\cite{miao2023specinfer,cai2023medusa,chen2023accelerating}.
While promising, the effectiveness of these techniques, especially when combined with autoregressive LLMs, remains largely unexplored.
This paper addresses two critical questions:
\textbf{\textit{Q1}}: Can existing linear attention methods, primarily designed for encoder-based LLMs like BERT~\cite{devlin2018bert} or Vision Transformers (ViTs)~\cite{vit}, be effectively applied to autoregressive decoder-based LLMs?
\textbf{\textit{Q2}}: Can linear attention and speculative decoding be seamlessly integrated to address both bottlenecks concurrently during LLM training and serving?

We conduct the first comprehensive empirical exploration to evaluate the efficacy of linearized autoregressive LLMs and their compatibility with speculative decoding.
Our findings for \textbf{\textit{Q1}} reveal that directly applying existing linear attention methods to autoregressive LLMs leads to suboptimal performance, due to the disruption of temporal dependencies crucial for autoregressive generation.
For instance, convolution-based augmentation techniques~\cite{you2023castling,xiong2021nystromformer} introduce ``\emph{information leakage}'' from future tokens during training, \hr{i.e., they use convoluted future context directly instead of predicting the next tokens}.
Addressing \textbf{\textit{Q2}}, we find that direct integration of linear attention with speculative decoding is ineffective, owing to mismatches in handling temporal dependencies. In particular, 
speculative decoding employs ``tree-based'' attention, complicating the application of standard linear attention methods.
\begin{figure}[t]

    \begin{minipage}{0.242\textwidth}
        \centering
        \includegraphics[width=\linewidth]{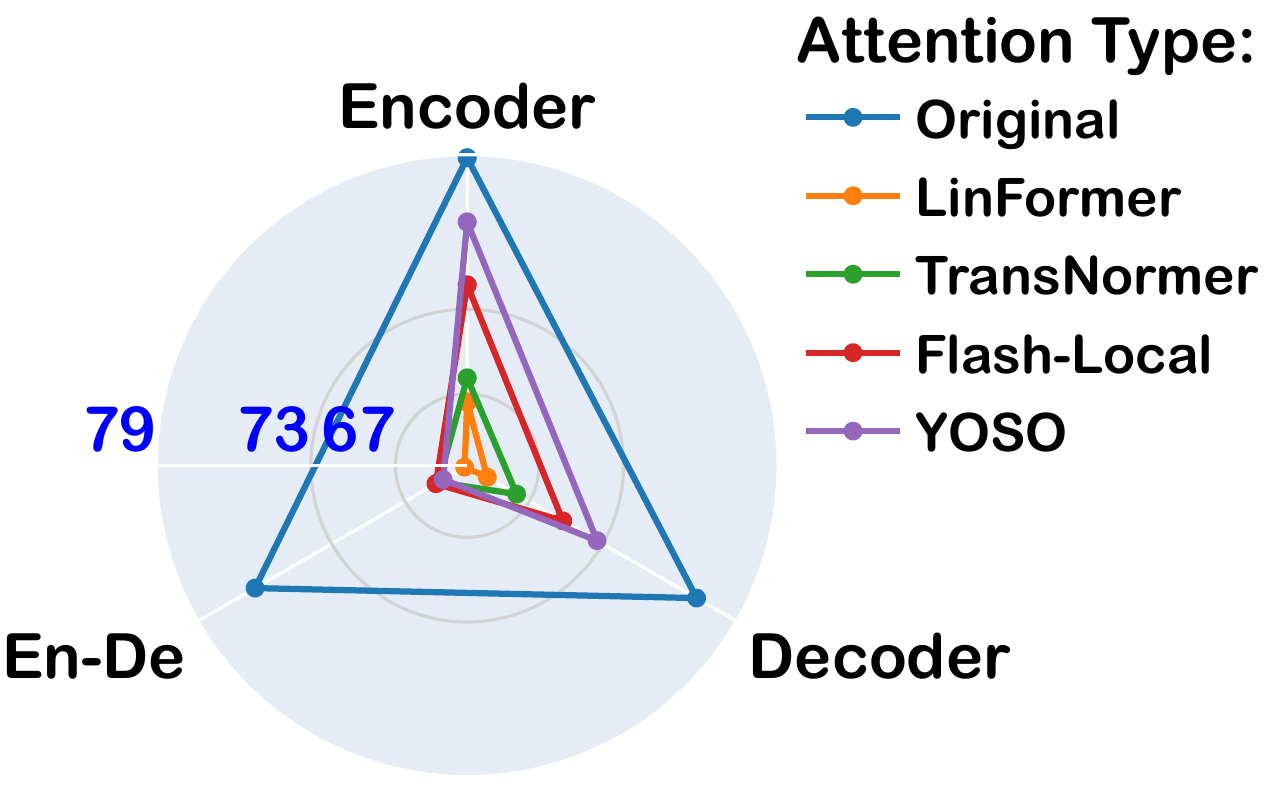}
    \end{minipage}
    \hspace{-0.22cm}
    \begin{minipage}{0.242\textwidth}
        \centering
        \includegraphics[width=\linewidth]{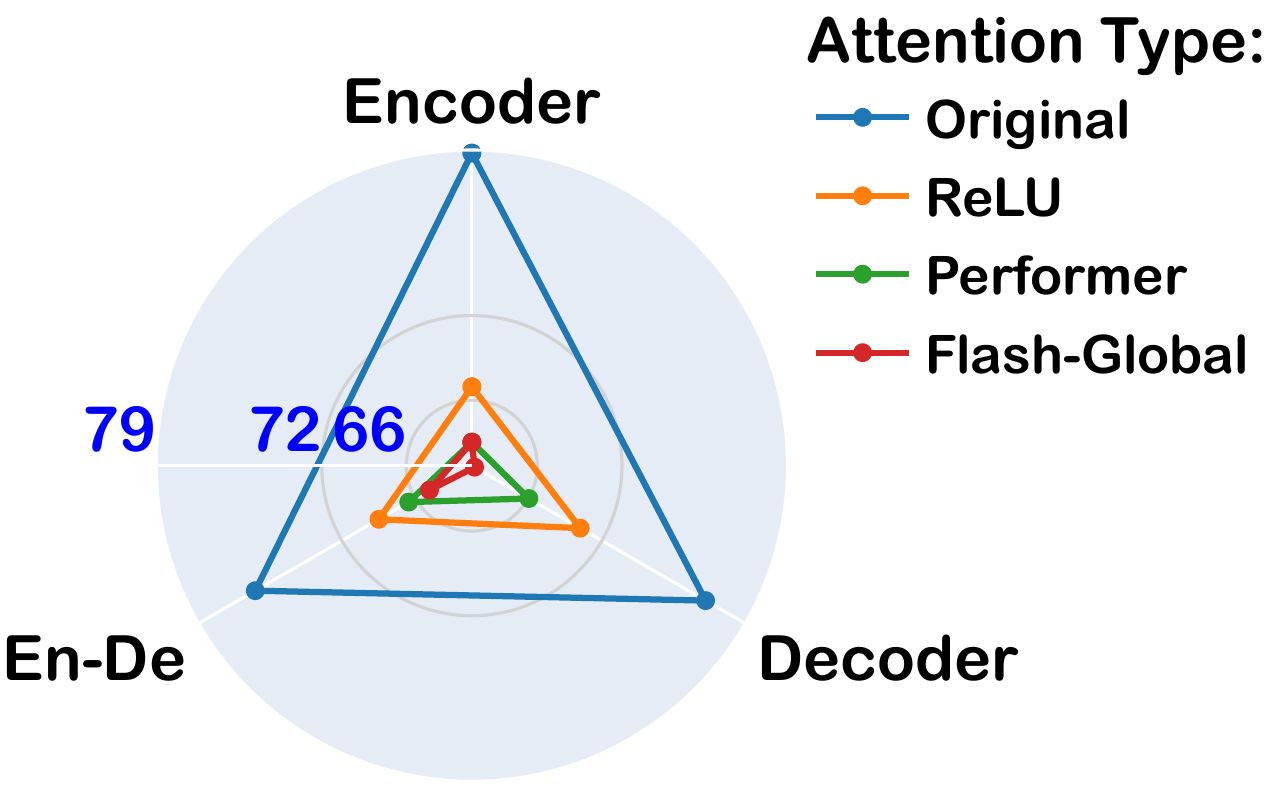}
    \end{minipage}
\caption{Empirical evaluation of seven linear attention methods on top of three types of LLMs on the GLUE~\cite{wang2018glue} benchmark: (1) encoder-based BERT~\cite{devlin2018bert}; (2) decoder-based GPT-2~\cite{gpt2}; and (3) encoder-decoder T5~\cite{roberts2022t5x}. \textbf{Left:} The majority of SOTA linear attentions, including LinFormer~\cite{wang2020linformer}, TransNormer\cite{transnormer}, FLASH-Local~\cite{flash}, and YOSO~\cite{yoso}, exhibit superior performance on encoder-based models compared to decoder-based ones. \textbf{Right:} Other linear attention methods, such as ReLU-based one~\cite{cai2023efficientvit}, Performer~\cite{performers}, and FLASH-Global~\cite{flash}, consistently perform less effectively on all LLMs.}
    \label{fig:exploration}
\end{figure}
Motivated by these challenges, we propose an effective local convolutional augmentation to prevent information leakage, boost performance, and maintain compatibility with speculative decoding. Our key contributions are:
\begin{itemize}[topsep=0pt, itemsep=0pt, leftmargin=1em]
    \item We conduct a comprehensive evaluation of seven linear attention methods across three types of LLMs (encoder-based, decoder-based, and encoder-decoder), revealing that existing encoder-based linear attentions are not optimally suited for autoregressive decoder-based LLMs.
    \item We introduce an \textbf{\textit{effective}} local augmentation technique that enhances the local feature extraction capabilities of linear attention in autoregressive LLMs while preventing information leakage.
    \item We develop a solution for seamlessly integrating linear attention with speculative decoding's tree-based attention, boosting token-level parallelism for \textbf{\textit{efficient}} generation and accelerating both LLM training and serving.
    \item Extensive experiments on five LLMs validate the effectiveness of our augmented linearized LLMs, achieving up to a 6.67 reduction in perplexity and up to 2$\times$ speedups during generation over existing linear attention methods.
    
\end{itemize}

\section{Related Works}
\label{sec:related_works}
\textbf{Autoregressive LLMs.}
Existing LLMs are broadly categorized into three architectures: \textit{encoder-based}, \textit{decoder-based}, and \textit{encoder-decoder} models.
Encoder-based models like BERT~\cite{devlin2018bert} focus on natural language understanding and are also commonly used in image processing~\cite{vit}. 
Encoder-decoder models, such as Transformer~\cite{transformer},
are designed for sequence-to-sequence tasks, where the encoder extracts features and the decoder generates outputs.
Decoder-based models, including GPT~\cite{gpt2,gpt4} and LLaMA~\cite{touvron2023llama}, generate text sequentially by predicting the next token.
While all these models utilize Transformer architectures, their specific design and purpose vary.
This paper presents a comprehensive study of applying linear attention techniques to both encoder-decoder and decoder-based LLMs.

\textbf{Efficient Linear Attention}
Self-attention in transformers, with their quadratic computational complexity~\cite{zhu2021long,katharopoulos2020transformers}, have led to the development of linear attention methods. 
Kernel-based linear attentions~\cite{liu2021swin,arar2022learned,wang2020linformer,tu2022maxvit} decompose the softmax with kernel functions and change the computation order.
However, few approaches focus on decoder-based autoregressive LLMs~\cite{flash,katharopoulos2020transformers}.
Recent studies, such as LongLoRA~\cite{chen2023longlora}, aim to adapt local attention techniques for efficient fine-tuning, but a thorough comparison of linear attention methods for autoregressive LLMs is less explored.
This paper systematically review existing linear attention for decoder-based autoregressive LLMs and investigates how to efficiently enhance less effective linear attention methods.

\textbf{Speculative Decoding.}
Linear attention methods reduce training inefficiencies, but the sequential nature of autoregressive decoding limits parallelism during deployment, restricting the number of input tokens.
Speculative decoding~\cite{chen2023accelerating,miao2023specinfer,kim2023big,leviathan2023fast,cai2023medusa} has proven to be an effective strategy for boosting parallelism in LLM serving. 
It utilizes small speculative models for initial generation, with the original LLMs validating the outputs.
Recent works, such as Medusa~\cite{cai2023medusa}, suggests that these models can be the same. This paper investigates the synergy between linearized LLMs and speculative sampling to improve LLM training and serving efficiency.

Additional related works are provided in Appendix~\ref{appendix:related_work}.

\section{Preliminaries and Evaluation}
\label{sec:evaluation}
\begin{figure}[t]
    \centering
    \includegraphics[width=\linewidth]{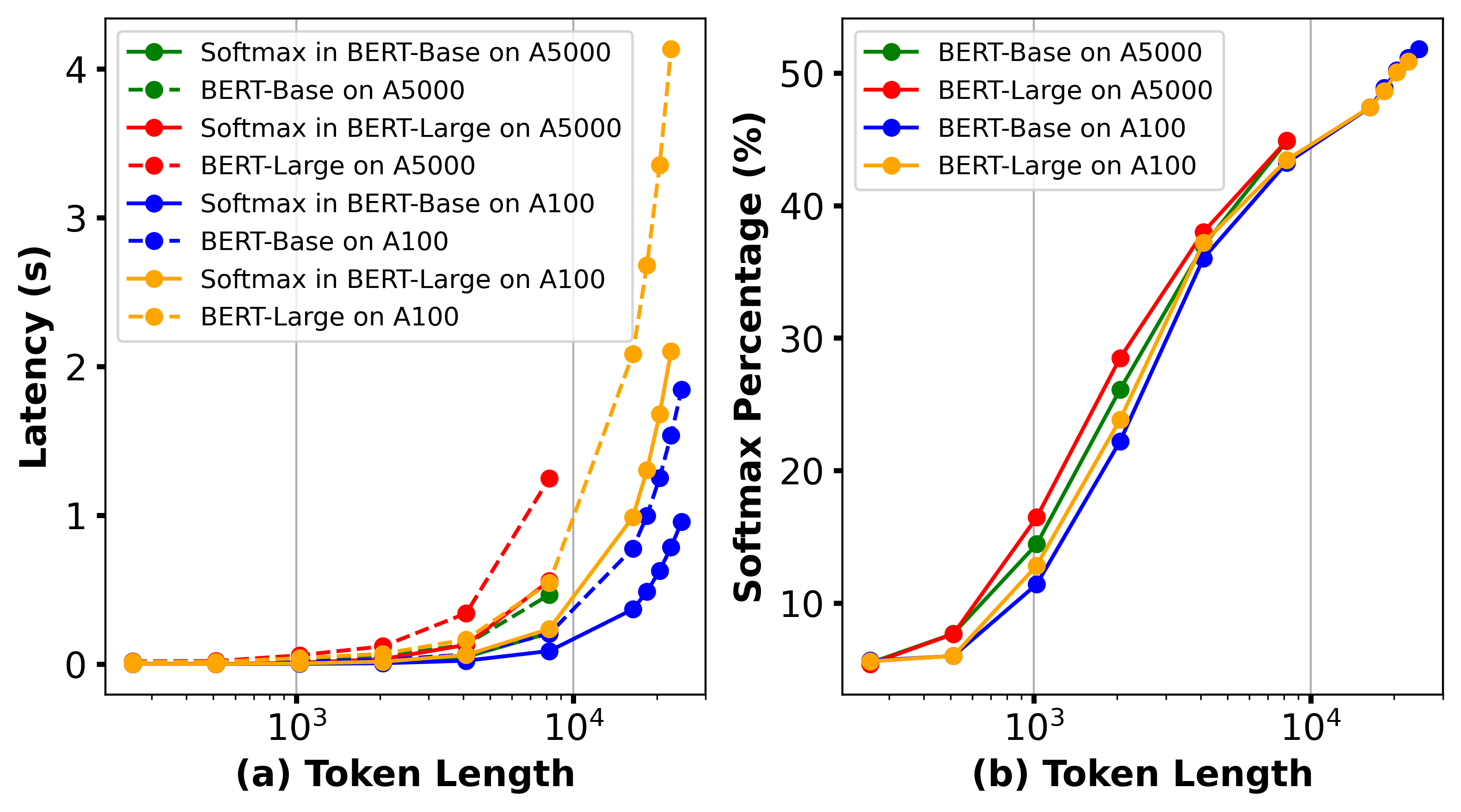}
    \caption{Runtime profiling: \textbf{(a)} actual runtime latencies for both the softmax and the entire model; \textbf{(b)} the percentage of time allocated to softmax computations across the latency of the entire model. All data were collected using BERT-Base/Large models on a single A5000 or A100 GPU.}
    \label{fig:softmax}
\end{figure}
\textbf{Self-Attention and Softmax Bottleneck.}
The self-attention module is a core component of the Transformer~\cite{transformer,vit}, and typically includes multiple heads.
Each head computes global-context information by evaluating pairwise correlations among all $n$ tokens ($n$ represents the total number of tokens) as follows:
\begin{equation} \label{equ:attn}
    \begin{split}
    & \texttt{\textbf{Attn}}(\mathbf{X}) = {\tt Concat}(\text{H}_1, \cdots, \text{H}_h) \cdot \mathbf{W}_O,
    \,  \mathrm{where} \\
    & \text{H}_i = {\tt Softmax}\left(\frac{f_Q(\mathbf{X}) \cdot f_K(\mathbf{X})^T}{\sqrt{d_k}}\right) \cdot f_V(\mathbf{X}),
    \end{split}
\end{equation}
where $h$ denotes the number of heads. Within each head $H_i$, input tokens $\mathbf{X} \in \mathbb{R}^{n \times d}$ of length $n$ and dimension $d$ will be linearly projected to query, key, and value matrices, i.e., $\mathbf{Q}, \mathbf{K}, \mathbf{V} \in \mathbb{R}^{n \times d_k}$, through three linear mapping functions, $f_Q=\mathbf{XW}_Q, f_K=\mathbf{XW}_K, f_V=\mathbf{XW}_V$, where $d_k = d / h$ is the dimensionality of each head and $\mathbf{W}_{Q/K/V}$ are the associated weight matrices.
The final outputs are generated by concatenating the results from all heads and applying a weight matrix $\mathbf{W}_O \in \mathbb{R}^{d \times d}$.

Attention is the bottleneck in LLMs, accounting for 55\% of the total runtime during autoregressive generation (Appendix \ref{sec:more_profiling}). 
Within self-attention, softmax becomes a memory bottleneck when handling long sequences~\cite{dao2022flashattention,flat}. 
As depicted in Fig. \ref{fig:softmax}, we profiled BERT-Base/Large models on a single A100/A5000 GPU to illustrate the percentage of time allocated to softmax as the token length increases. 
We observe that the runtime percentage for softmax continues to increase \textit{quadratically} as the token length grows, occupying \textit{up to 50\%} of the total model latency when token length reaches $10^4$.

\textbf{Linear Attentions (LAs).}
Kernel-based LAs~\cite{katharopoulos2020transformers,wang2020linformer,you2023castling} have emerged as an effective method for eliminating the need for softmax and reducing the quadratic complexity. The core idea is to decompose the similarity measurement function, typically based on softmax, into separate kernel embeddings, i.e., $\text{Sim}(\mathbf{Q}, \mathbf{K}) \approx \phi(\mathbf{Q}) \cdot \phi(\mathbf{K})^T$. This enables rearranging the computation order to $\phi(\mathbf{Q}) (\phi(\mathbf{K})^T \mathbf{V})$, utilizing the associative property of matrix multiplication. Consequently, the complexity of attention becomes linear relative to the feature dimension $d$, instead of quadratic with respect to the token length $n$. These LAs, however, could lead to a significant accuracy drop compared to softmax-based attention unless they are carefully designed.
\begin{figure}[t]
    \centering
    \includegraphics[width=\linewidth]{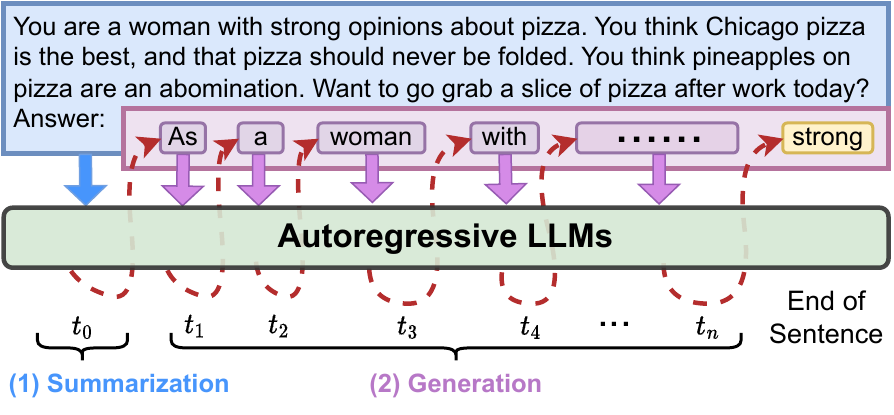}
    \caption{Illustrating the autoregressive LLMs. The process of generating text unfolds in two stages: (1) an initial \textit{summarization or prefill phase} that employs a large batch size and utilizes the given input context; followed by (2) the \textit{generation or decode phase}, which operates on a single-batch basis, using previously generated tokens to continue the text output.
    }
    \label{fig:autoregressive_llm}
\end{figure}

\textbf{Autoregressive LLMs.}
As depicted in Fig. \ref{fig:autoregressive_llm}, unlike the initial summarization phase, which processes a large number of tokens simultaneously and is computationally intensive, the generation phase faces severe memory or bandwidth limitations due to its autoregressive nature, involving token-by-token generation. 
Linear attention speeds up training and reduces summarization complexity, but is less effective for autoregressive generation due to low parallelism.
Speculative decoding emerges as a critical method for increasing parallelism. Thus, ensuring compatibility between linear attention and speculative decoding is imperative for efficient summarization and generation phases.

\subsection{Evaluation of Existing LAs on LLMs}\label{sec:eval-exist}
\begin{table}[t]
\setlength\tabcolsep{3pt}
\caption{
Evaluation of seven LAs on BERT~\cite{devlin2018bert}, an encoder-based LLM, with the text classification accuracy on the GLUE benchmark~\cite{wang2018glue}.
}
\label{tab:bert}
\begin{center}
\renewcommand{\arraystretch}{1.1} %
\resizebox{\linewidth}{!}{
\begin{tabular}{l|ccccccc|c}
\Xhline{3\arrayrulewidth}
\textbf{BERT w/ LAs} & \textbf{SST2} & \textbf{WNLI} & \textbf{QNLI} & \textbf{MNLI} & \textbf{RTE} & \textbf{MRPC} & \textbf{QQP} & \textbf{Average}\\
\hline
 \textbf{BERT} (Baseline)  & 93.58	& 42.25&91.49& 84.81& 66.43&83.09&91.10&\textbf{78.96} \\
\Xhline{3\arrayrulewidth}
\textbf{FLASH-Local}    & 91.63	&47.89&	88.38	&	81.06	&50.18	&70.10	&90.56	&\textbf{74.26}    \\
\textbf{FLASH-Global}     & 76.72	&54.93	&53.69	&33.46&	48.74	&68.63&	78.32	&\textbf{59.21}\\
\textbf{Linformer} & 81.54&	56.34&63.06&67.54&45.13&68.38	&81.32&	\textbf{66.19}
\\
\textbf{Performer}   & 80.16	&45.07	&60.77	&39.81	&45.49	&67.40	&75.88	&\textbf{59.23} \\
\textbf{TransNormer}    & 81.88	&56.34	&67.67	&67.01	&53.07	&70.10	&83.13	&\textbf{68.46} \\
\textbf{YOSO}      & 91.51	&52.11	&87.75	&82.16&	58.12	&75.98	&90.40 &	\textbf{76.86}
\\
\textbf{ReLU}      & 81.77	&56.34&	61.54&	70.14&	47.29	& 69.85&	82.44	&\textbf{67.05}
\\
\Xhline{3\arrayrulewidth}
\end{tabular}
}
\end{center}
\end{table}

\begin{table}[t]
\setlength\tabcolsep{3pt}
\caption{
Evaluation of seven LAs on GPT-2~\cite{gpt2}, a decoder-based LLM, with the text classification accuracy on the GLUE benchmark~\cite{wang2018glue}.
}
\label{tab:gpt}
\begin{center}
\renewcommand{\arraystretch}{1.1} %
\resizebox{\linewidth}{!}{
\begin{tabular}{l|ccccccc|c}
\Xhline{3\arrayrulewidth}
\textbf{GPT-2 w/ LAs} & \textbf{SST2} & \textbf{WNLI} & \textbf{QNLI} & \textbf{MNLI} & \textbf{RTE} & \textbf{MRPC} & \textbf{QQP} & \textbf{Average}\\
\hline
GPT-2 (Baseline)  & 91.28	& 57.75	& 88.39	&81.54	& 60.65&	74.51	& 89.13	& \textbf{77.61} \\
\Xhline{3\arrayrulewidth}
\textbf{FLASH-Local}    & 83.60&	53.52	& 77.16	& 73.97	&48.01	& 68.87	& 86.40&	\textbf{70.22}   \\
\textbf{FLASH-Global}     & 50.92	& 50.70&	54.27	& 34.59	& 52.35	& 68.38	& 63.19	& \textbf{53.49}\\
\textbf{Linformer} & 79.47	& 52.11&	60.96	& 34.56	& 52.35	& 68.38	& 76.30	& \textbf{60.59}
\\
\textbf{Performer}   & 86.93	& 38.03	& 69.36& 70.60&	49.46	&69.12	&76.30&	\textbf{65.69} \\
\textbf{TransNormer}    & 82.11	&56.34	&63.48	&59.11	&53.07	&68.38	&75.79	& \textbf{65.47} \\
\textbf{YOSO}      & 88.42	&45.07&	82.23&	77.80&	54.51	&73.04&	87.72&	\textbf{72.68}
\\
\textbf{ReLU}      & 86.47&	45.07	&80.96	&78.02	&51.99&	69.61	&83.42	&\textbf{70.79}
\\
\Xhline{3\arrayrulewidth}
\end{tabular}
}
\end{center}
\end{table}

\begin{table}[t]
\setlength\tabcolsep{3pt}
\caption{
Evaluation of seven LAs on T5~\cite{2020t5}, an encoder-decoder-based LLM, with the text classification accuracy on the GLUE benchmark~\cite{wang2018glue}.
}
\label{tab:t5}
\begin{center}
\renewcommand{\arraystretch}{1.1} %
\resizebox{\linewidth}{!}{
\begin{tabular}{l|ccccccc|c}
\Xhline{3\arrayrulewidth}
\textbf{T5 w/ LAs} & \textbf{SST2} & \textbf{WNLI} & \textbf{QNLI} & \textbf{MNLI} & \textbf{RTE} & \textbf{MRPC} & \textbf{QQP} & \textbf{Average}\\
\hline
\textbf{T5} (Baseline)  & 93.81	& 36.62	& 91.73	& 86.54	&58.12&	80.64	& 90.89	& \textbf{76.91} \\
\Xhline{3\arrayrulewidth}
\textbf{FLASH-Local}    &  77.87	& 56.34	& 58.87 & 49.44&	52.71	& 68.38	&75.62	& \textbf{62.75}   \\
\textbf{FLASH-Global}     & 80.62&	56.34	& 63.65	& 49.87&	46.93	&68.38	&79.29	& \textbf{63.58}\\
\textbf{Linformer} & 51.15	&43.66	& 55.43		& 46.60	& 51.99	&68.38	&74.19	& \textbf{55.91}
\\
\textbf{Performer}   & 82.57	& 56.34	& 63.70		& 61.75	& 52.35	& 69.85	& 78.60	& \textbf{66.45} \\
\textbf{TransNormer}    & 
79.36	& 43.66	& 59.78		& 48.75	& 58.48	& 70.59	& 75.37& \textbf{62.29}\\
\textbf{YOSO}      &  78.33	& 56.34&	59.55	&48.64	& 47.65	&68.38&	70.87	& \textbf{61.39}
\\
\textbf{ReLU}      & 85.79	& 53.52	& 71.57	& 73.52&	48.01	&70.34	& 83.89	& \textbf{69.52}
\\
\Xhline{3\arrayrulewidth}
\end{tabular}
}
\end{center}
\end{table}

\textbf{Comprehensive Evaluation.}
To investigate whether previous LAs can be generally applicable to three categories of LLMs: encoder-based, decoder-based, and encoder-decoder, we evaluate seven distinct LAs, including FLASH-Local\&Gloabl~\cite{flash}, Linformer~\cite{wang2020linformer}, Performer~\cite{performers}, TransNormer~\cite{transnormer}, YOSO~\cite{yoso}, ReLU~\cite{cai2023efficientvit}, across three representative LLMs in each category: encoder-based BERT~\cite{devlin2018bert}, decoder-based GPT-2~\cite{gpt2}, and encoder-decoder T5~\cite{2020t5}. 
As detailed in Tabs. \ref{tab:bert}, \ref{tab:gpt}, and \ref{tab:t5}, we have applied these LAs to their respective LLMs, assessing their performance across seven linguistic tasks from the General Language Understanding Evaluation (GLUE) benchmark~\cite{wang2018glue}.
To enhance comparison efficacy, we also report the accuracy of softmax-based LLMs as a baseline. This facilitates a straightforward evaluation of the average accuracy drop across the seven LAs and the seven tasks when being applied to different types of LLMs.

\textbf{Result Analysis.}
Our evaluation shows that:
(1) most LAs are effective in encoder-based LLMs, aligning with their initial design intent. However, their performance diminishes when applied to decoder-based or encoder-decoder-based LLMs. 
On average, seven LAs applied to encoder-based LLMs result in an average accuracy of 67.32, whereas for decoder-based or encoder-decoder-based models, the accuracy drops to 65.56 and 63.13, respectively;
(2) as shown in Fig. \ref{fig:exploration} (left), advanced LA techniques perform well in encoder-based LLMs but struggle to replicate these results in decoder or encoder-decoder-based LLMs.
For instance, FLASH-Local~\cite{flash} and YOSO~\cite{yoso} achieve score 74.26/76.86 on BERT, only slightly below the baseline, but drops to 70.22/72.68 on GPT-2 and further to 62.75/61.39 on T5, significantly lower than their softmax-based counterparts;
(3) as shown in Fig. \ref{fig:exploration} (right), LAs that are less effective in encoder-based LLMs consistently underperform in decoder-based and encoder-decoder based LLMs, highlighting their distinct suitability for different LLM architectures.

\begin{figure}[t]
    \centering
    \includegraphics[width=\linewidth, height=0.53\linewidth]{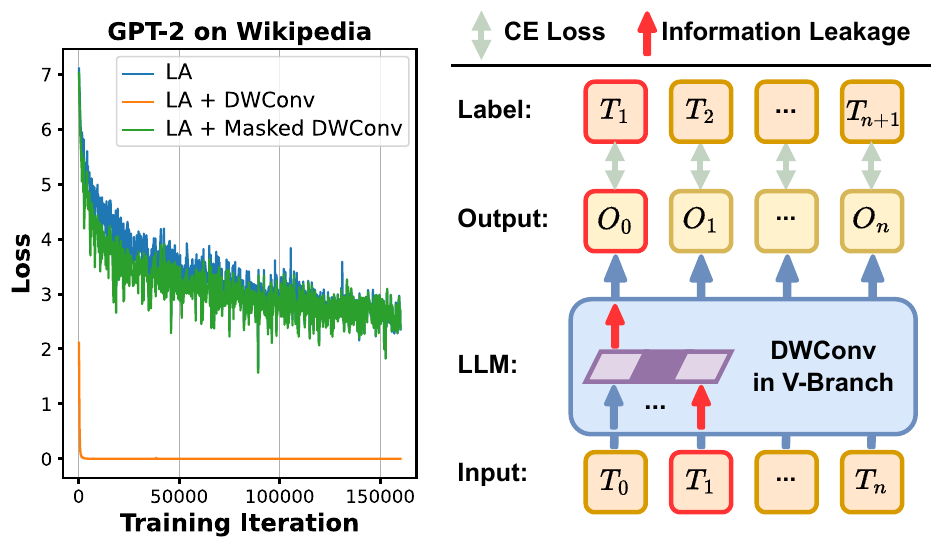}
    \caption{
    Existing augmented LAs fail in autoregressive LLMs. \textbf{Left}: The augmented DWConv branch results in zero loss/accuracy, as indicated by the yellow line. \textbf{Right}: Illustration of the information leakage phenomenon, i.e., next tokens are prematurely revealed as shown by red arrows, in autoregressive LLMs with DWConv in the $\mathbf{V}$ branch.
    }
    \label{fig:la_conv}
\end{figure}

\textbf{Limitations of Existing LAs.}
Our evaluation indicates that most LAs suffer an accuracy drop in autoregressive decoder-based LLMs in generation tasks. 
Advanced LA techniques, such as efficient depthwise convolution (DWConv) in the $\mathbf{V}$ (value) branch of attention modules~\cite{you2023castling,xiong2021nystromformer},  fail in autoregressive LLMs due to an information leakage from the inclusion of future context during training. 
As evident in Fig. \ref{fig:la_conv}, LA with DWConv convergences to zero loss early in training, but actual evaluation accuracy remains zero, indicating information leakage as also depicted in Fig. \ref{fig:la_conv} (right).
In addition, while LAs improve training and summarization, their effectiveness is limited in token-by-token generation and compatibility with speculative decoding to increase parallelism during generation remains challenging. We will further discuss our augmented methods for autoregressive LLMs and their integration with speculative decoding in subsequent sections.
\section{The Proposed Method}
\label{sec:methods}
In this section, we introduce a revised local augmentation technique for existing LAs to enhance accuracy and examine the synergy of augmented LAs with speculative decoding for both efficient LLM training and autoregressive generation.

\begin{figure}[t]
    \centering
    \includegraphics[width=0.9\linewidth]{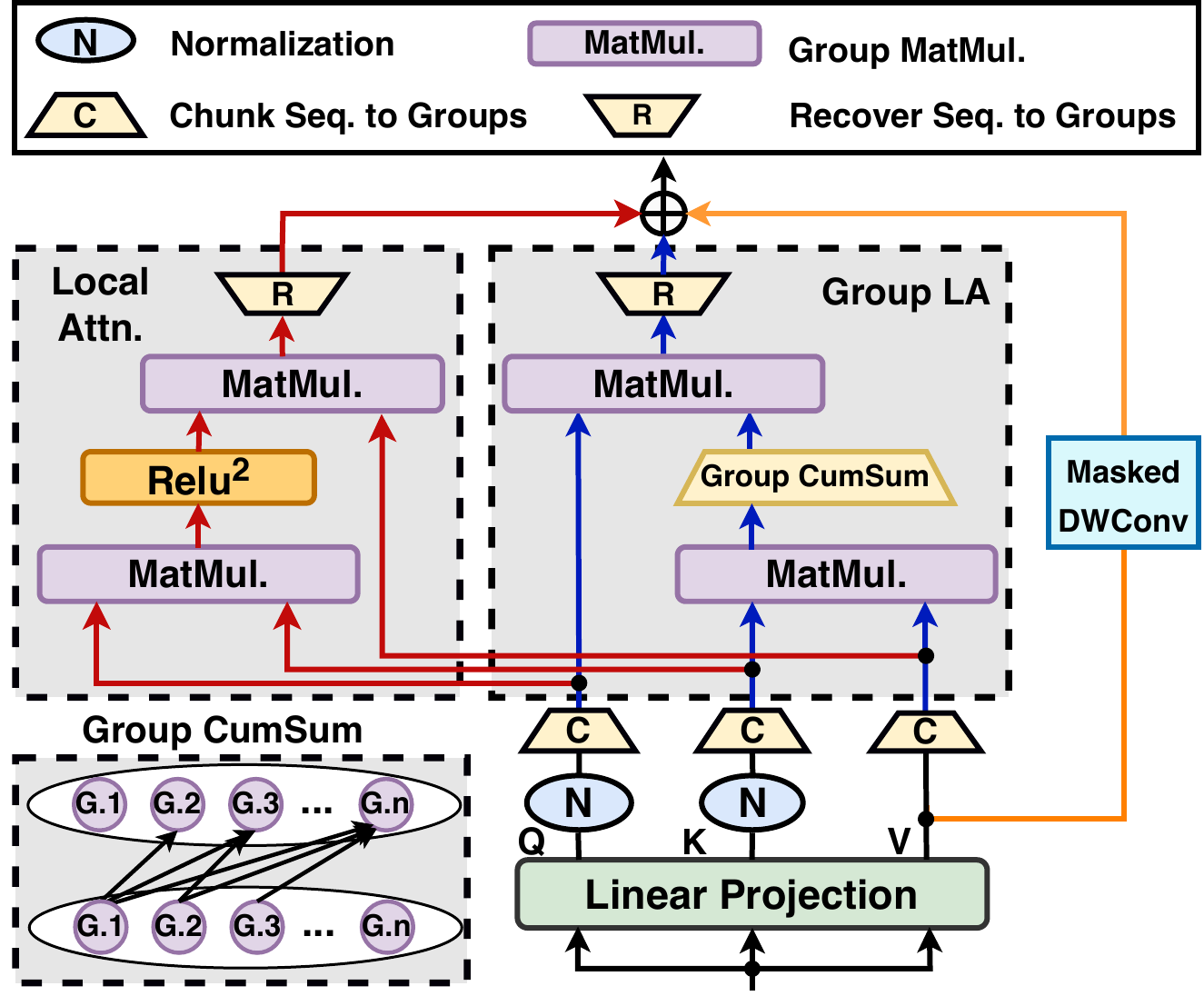}
    \caption{Model architecture of our LA augmentation.
    }
    \label{fig:la_aug}
\end{figure}

\subsection{LA Augmentation for Autoregressive LLMs}

\textbf{Revised LA Augmentation.} 
To address information leakage, we propose to design an effective masked DWConv instead of using a simple convolutional layer for enhancing the locality of the linear attention~\cite{you2023castling,xiong2021nystromformer}. 
Specifically, we adopt a causal mask on the DWConv layer to prevent tokens from accessing information from subsequent tokens, thereby preserving the inherent causality of the original attention mechanism, as illustrated in the right branch of Fig. \ref{fig:la_aug}.
The masked DWConv prevents information leakage, leading to better loss convergence, as demonstrated in the left of Fig. \ref{fig:la_conv}.
Unlike \cite{dauphin2017language}, our efficient DWConv is integrated directly into the attention block, not as a standalone component.

We build our DWConv augmentation on top of existing grouped LAs to speed up the linearized LLMs. The reason why we need the grouped LA is that standard LAs exhibit reduced efficiency in autoregressive settings due to the causal constraint~\cite{flash}. 
For example, the query vector $\mathbf{Q_t}$ at $t$-th time step interacts with the cumulative sum of all preceding results $\sum_{i=1}^{t} \mathbf{K_i} \mathbf{V_i}$. 
This cumulative sum (\texttt{cumsum}) of $\mathbf{KV}$ product operations inherently creates a sequential dependency, and restricts the potential for parallel processing.
To enhance efficiency, we partition the input sentence into non-overlapping groups. Within each group, we bypass local dependencies, allowing parallel processing. For interactions between groups, we only compute the cumulative sums at the group level for the $\mathbf{KV}$ products for improved efficiency, as depicted in the middle branch of Fig. \ref{fig:la_aug}.
Furthermore, to improve local dependency handling, we employ parallel local attention within each group, using softmax-based attention, as depicted in the left branch of Fig. \ref{fig:la_aug}. The integration of this local attention strategy with our revised local augmentation contributes significantly to the performance, combining the efficiency of LAs with improved accuracy. 
\begin{figure}[t]
    \centering
    \includegraphics[width=\linewidth]{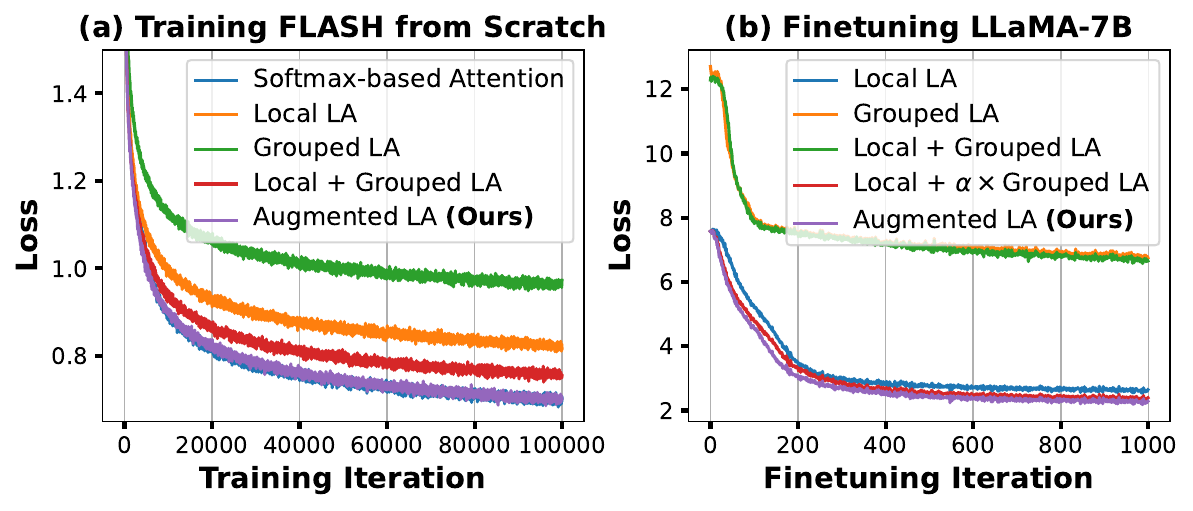}
    \caption{Tested our augmented linear attention mechanism for both training from scratch and fine-tuning from pre-trained model settings, where \textbf{(a)} shows the training progress of FLASH models~\cite{flash}; \textbf{(b)} depicts the fine-tuning performance of LLaMA-7B~\cite{touvron2023llama}.}
    \label{fig:flash_llama_loss}
\end{figure}

\textbf{Verification on Small- and Large-Scale LLMs.}
We evaluate and verify the revised LA augmentation on both small- and large-scale LLMs, i.e., FLASH~\cite{flash} and LLaMA-7B~\cite{touvron2023llama}.
For FLASH, we train a small model from scratch for 100K steps on enwik8~\cite{hutter2012human}. As shown in Fig. \ref{fig:flash_llama_loss} (a), grouped LA leads to reduced accuracy or increased loss.
Local LA alone is also ineffective. 
A combination of grouped and local LAs showed some improvement but remained inferior to the traditional softmax-based attention method. 
In contrast, our augmented LAs, blending the grouped LA concept with masked DWConv augmentation (with a kernel size of 63), achieved the most favorable results among all LAs, on par with the original softmax-based attentions.
For LLama-7B, we finetune it using LAs on the RedPajama dataset~\cite{together2023redpajama} for 1K steps with a batch size of 64 following~\cite{chen2023longlora}.
Fig. \ref{fig:flash_llama_loss} (b) indicates a similar trend to FLASH, where local augmentation proves even more vital in this finetuning phase, and reliance solely on global LA leads to significantly higher loss.
Note that we use a hyperparameter $\alpha$ to balance the interplay between global and local LAs. Overall, our augmented LAs combining the three branches in Fig. \ref{fig:la_aug} consistently outperform existing LAs.
\subsection{When LA Meets Speculative Decoding}
To address limited parallellism in LLM serving, we aim to combine speculative decoding with our imporved LAs.
However, direct integration is ineffective. Here, we explore compatibility challenges and propose seamless solutions.
\begin{figure}[t]
    \centering
    \includegraphics[width=\linewidth]{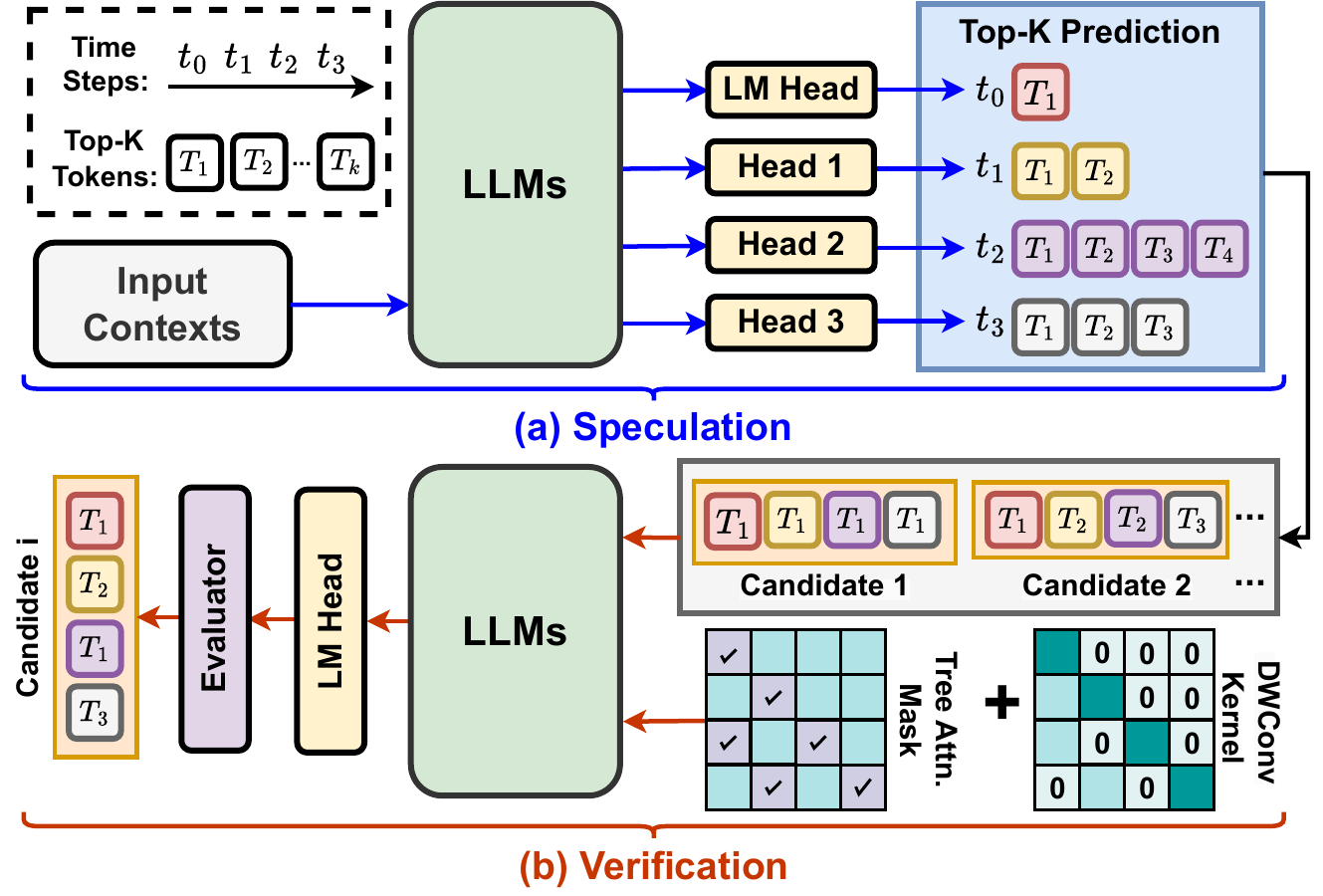}
    \caption{Illustrating the speculative decoding pipeline with our augmented LAs: (a) Speculation; and (b) Verification.}
    \label{fig:speculative_decoding}
\end{figure}

\begin{figure}[t]
    \centering
    \includegraphics[width=\linewidth]{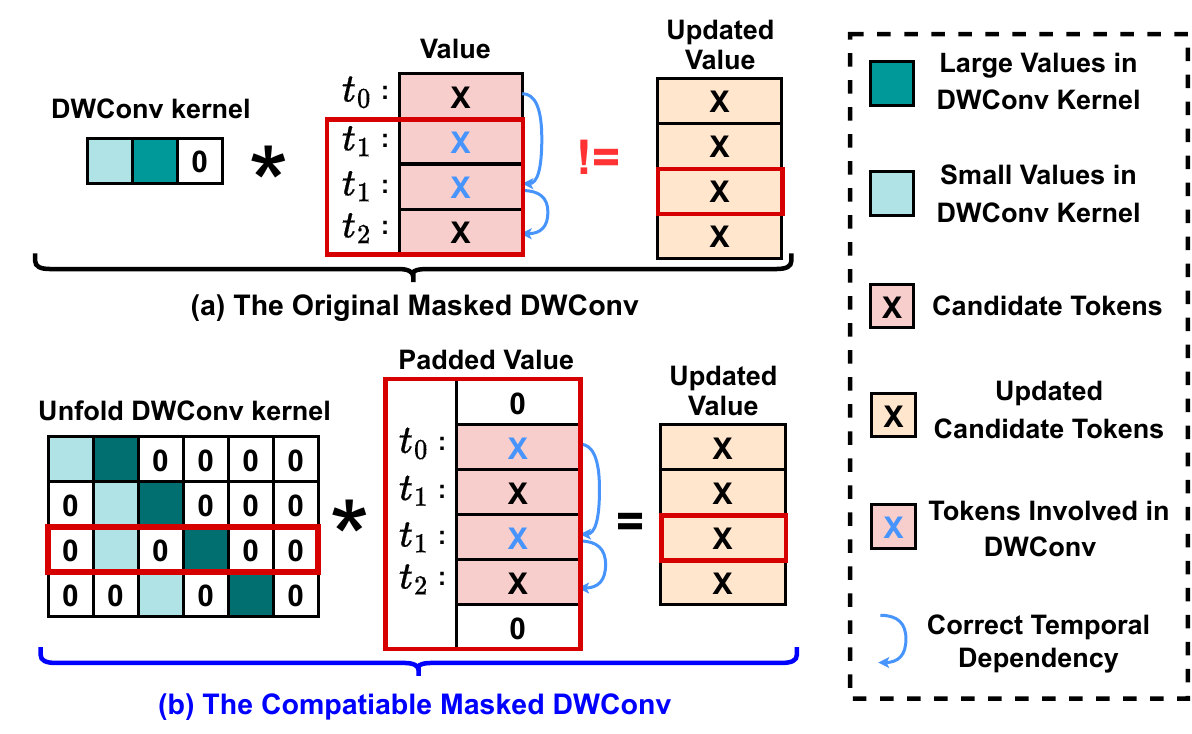}
    \caption{\textbf{(a)}: DWConv itself fails to capture the temporal dependency in speculative decoding; \textbf{(b)}: Our Unfolded DWConv kernels capture the correct temporal dependency.}
    \label{fig:DWConv}
\end{figure}

\textbf{Compatibility Analysis.}
Speculative decoding, such as Medusa~\cite{cai2023medusa}, uses smaller draft models to simultaneously predict multiple output tokens across different time steps, as illustrated in Fig. \ref{fig:speculative_decoding} (a).
The original LLMs then act as verifiers, either accepting or rejecting them, and resampling if needed, as illustrated in Fig. \ref{fig:speculative_decoding} (b). 
This approach improves parallelism during LLM generation.
However, combining LAs with speculative decoding is challenging because it generates multiple candidate outputs per step, with varying counts per time step, altering the temporal dependency.
This change is not effectively captured by masked DWConvs and grouped LAs in our augmented LAs.
As shown in Fig. \ref{fig:DWConv} (a), using a masked DWConv with a kernel size of 3 to convolve over stacked candidate tokens at time step $t_1$ results in capturing time steps $\{t_1, t_1\}$, rather than the correct sequence $\{t_0, t_1\}$. This discrepancy occurs because, at time step $t_1$, two candidate tokens are included in the convolution instead of the final verified one, leading to a temporal misalignment.

\textbf{Proposed Solution.}
To integrate our augmented LAs with the speculative decoding, we propose the updated design of DWConv and grouped LA to take into consideration the temporal dependencies represented in Medusa's tree-based attention mask. 
This design ensures the simultaneous processing of multiple candidate tokens while ensuring that each token only accesses information from its preceding token.
As shown in Fig. \ref{fig:DWConv} (b), we unfold the convolution into matrix multiplication, akin to the img2col method~\cite{vasudevan2017parallel}. 
This unfolding allows for the integration of tree-based attention masks with DWConv kernels, addressing their compatibility with negligible overheads.
For example, using a masked DWConv with an unfolded kernel to convolve over stacked candidate tokens at time step $t_1$ successfully captures the correct sequence $\{t_0, t_1\}$, while omitting an unchosen candidate at the same time step $t_1$.
In addition, we categorize speculative tokens into groups based on temporal dependency, regardless of the number of candidates per time step. In this way, tokens in each group interact only with verified tokens from previous groups, aligning their visibility with the tree-based attention pattern.
\begin{figure*}[t]
    \centering
    \includegraphics[width=\linewidth]{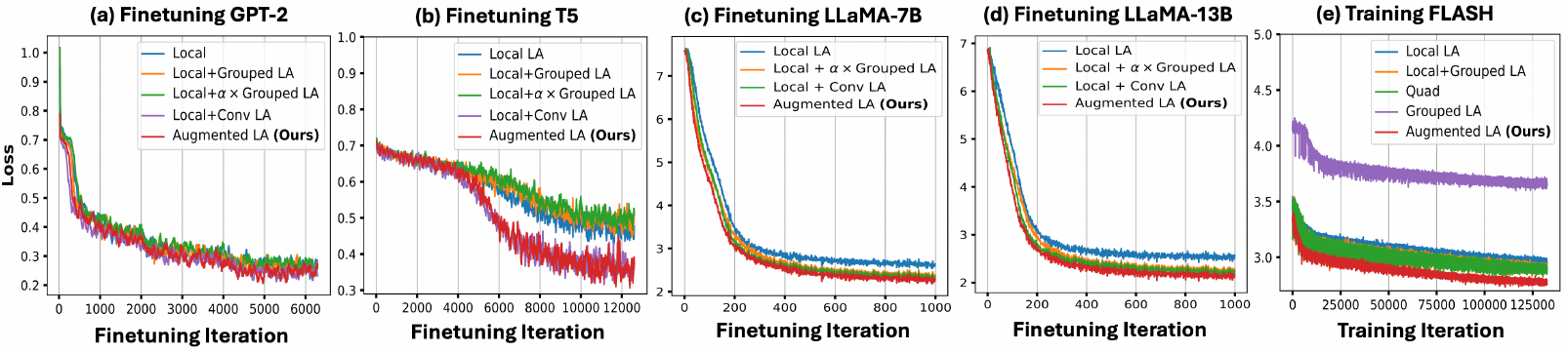}
    \caption{Visualizing the training trajectories of baseline LAs and our augmented LAs.}
    \label{fig:training_loss}
\end{figure*}
\begin{table*}[t]
\setlength\tabcolsep{4pt}
\caption{Inference latency and memory comparison at various sequence lengths for LLaMA models.}
\label{tab:overall_comp}
\begin{center}
\renewcommand{\arraystretch}{1.1} %
\resizebox{\linewidth}{!}{
\begin{tabular}{l|c|ccccc|ccccc}
\Xhline{3\arrayrulewidth}
\multirow{2}{*}{\textbf{Model}} & \multirow{2}{*}{\textbf{Attn.}} & \multicolumn{5}{c|}{\textbf{Inference Latency (ms)}} & \multicolumn{5}{c}{\textbf{Inference Memory (GB)}} \\
\cline{3-12}
 & & \textbf{2048} & \textbf{4096} & \textbf{8192} & \textbf{16384} & \textbf{32768} & \textbf{2048} & \textbf{4096} & \textbf{8192} & \textbf{16384} & \textbf{32768} \\
\Xhline{3\arrayrulewidth}
\multirow{2}{*}{LLaMA-2-7B} & Original & 302.3 & 812.6 & 2355.0 & OOM & OOM & 16.6 & 22.5 & 40.5 & OOM & OOM \\ 
 & \textbf{Ours LA} & 275.2 \textbf{(-9\%)} & 529.8 \textbf{(-35\%)} & 1029.7 \textbf{(-56\%)} & 2032.9 & 3985.9 & 15.9 \textbf{(-4\%)} & 19.1 \textbf{(-15\%)} & 25.7 \textbf{(-37\%)} & 38.8 & 65.0 \\ \hline
\multirow{2}{*}{LLaMA-2-13B} & Original & 491.6 & 1319.7 & 3805.0 & OOM & OOM & 30.1 & 38.2 & 62.1 & OOM & OOM \\ 
 & \textbf{Ours LA} & 449.4 \textbf{(-9\%)} & 876.5 \textbf{(-34\%)} & 1737.1 \textbf{(-54\%)} & 3460.9 & OOM & 29.1 \textbf{(-3\%)} & 33.8 \textbf{(-12\%)} & 43.2 \textbf{(-30\%)} & 61.9 & OOM \\ 
\Xhline{3\arrayrulewidth}
\end{tabular}
}
\end{center}
\end{table*}

\section{Experiments}
\label{sec:experiments}
\textbf{Models, Tasks, and Datasets.} \textit{\underline{Models.}} We apply our proposed augmented LA on top of five models, including FLASH~\cite{flash}, T5~\cite{2020t5}, GPT-2~\cite{gpt2}, LLaMA-2-7B, and LLaMA-2-13B~\cite{touvron2023llama2}.
In particular, we train the FLASH~\cite{flash} model of roughly 110M parameters from scratch and finetune the remaining language models of different sizes with our augmented LAs.
\textit{\underline{Tasks and Datasets.}} 
For FLASH and LLaMA-2-7B/13B models, we evaluate them on language modeling tasks.
Specifically, we train the FLASH model on the English partition of Wiki40b~\cite{guo-etal-2020-wiki}, which includes about 40B characters from 19.5M pages obtained from Wikipedia. 
We finetune the LLaMA-2-7B/13B models on RedPajama~\cite{together2023redpajama} dataset with about 1.2T tokens for 1K steps, following the setting of LongLora~\cite{chen2023longlora}. 
For T5 and GPT-2 models, we consider the text classification task to evaluate our augmented LAs, and choose seven datasets from GLUE~\cite{wang2018glue} benchmark: SST2~\cite{socher-etal-2013-recursive}, WNLI~\cite{wnli}, QNLI~\cite{rajpurkar-etal-2016-squad}, MNLI~\cite{williams-etal-2018-broad}, RTE~\cite{RTE}, MRPC~\cite{dolan-brockett-2005-automatically}, and QQP~\cite{Chen2017QuoraQP}.
In addition, we consider the evaluation of LLaMA models on six zero-shot or few-shot downstream tasks: BBH~\cite{suzgun2022challenging}, PIQA~\cite{bisk2020piqa}, MMLU~\cite{hendrycks2020measuring}, COPA~\cite{wang2019superglue}, ARCC~\cite{clark2018think}, and AGNews~\cite{zhang2015character}.
Following common evaluation settings, MMLU was tested with 5 shots, BBH with 3 shots, and the remaining tasks with zero shots.

\textbf{Training Settings.} 
\textit{\underline{For the FLASH training task}}, we train the model of roughly 110M parameters from scratch with a sequence length of 1024. The batch size is 256 and the token per batch is set to $2^{18}$. We use the AdamW optimizer with linear learning rate decay and a peak learning rate of $7\times10^{-4}$, the momentum of the AdamW optimizer is set to $\beta_1=0.9$, $\beta_2=0.95$ and the group size is set to 256 following~\cite{flash}. 
\textit{\underline{For the LLaMA-2 finetune task}}, we train it for 1K steps with a peak learning rate of $2\times10^{-5}$ and a batch size of 64. The learning rate scheduler is constant with 20 warmup steps. The optimizer is AdamW with the momentum of $\beta_1=0.9$ and $\beta_2=0.95$. The group size is set to 64 following~\cite{chen2023longlora}. \textit{\underline{For the GLUE task}}, We finetune the models for 3 epochs with a learning rate of $2\times 10^{-5}$ and a batch size of 32~\cite{devlin2018bert}. The group size is set to 64, and the sequence length is set to 256. 

\textbf{Baseline and Evaluation Metrics.} 
\textit{\underline{Baselines}}. 
For the text classification task on the GLUE benchmark, we compare the proposed augmented LAs with FLASH-Local\&Global~\cite{flash}, Linformer~\cite{wang2020linformer}, Performer~\cite{performers}, TransNormer~\cite{transnormer}, YOSO~\cite{yoso}, ReLU~\cite{cai2023efficientvit}. 
For the LLaMA-2 finetune tasks, we compare the proposed augmented LAs with the local and global attention proposed in~\cite{flash}, i.e., FLASH-Local/Global. 
For the FLASH training task, we compare our proposed method with local, global, and quadratic softmax-based attention. 
\textit{\underline{Evaluation Metrics}}. 
For the GLUE task, we use the classification accuracy to evaluate the augmented LA and baselines. 
For the LLaMA-2 finetune task,
we use the perplexity on PG-19~\cite{rae2019compressive} to evaluate all methods.
For the FLASH training task,
we use the validation set perplexity of Wiki40B to evaluate.
In addition, to evaluate the speedups after integrating our LAs and speculative decoding, we test the decoding speeds on MT-Bench~\cite{zheng2023judging} following~\cite{cai2023medusa}. 

\subsection{Our Linearized LLMs over Original LLMs}\label{sec:exp-main1}
We analyze the latency improvements and memory efficiencies of our linearized LLMs compared to conventional LLMs.
As detailed Tab.~\ref{tab:overall_comp}, our approach reduces latency by up to 56.3\% and memory usage by 36.5\% for models like  LLaMA-2-7B and LLaMA-2-13B on A100-80G device.
In addition, our linearized LLMs extend the maximum supported sequence lengths from 8K to 32K for LLaMA-2-7B on the same GPU, demonstrating our method's efficacy and scalability in large-scale models.
\begin{table}[t]
\setlength\tabcolsep{4pt}
\caption{Inference latency and memory comparison at various task prefill and decode sizes for LLaMA-2-7B models.}
\label{tab:overall_comp_2}
\begin{center}
\setlength{\tabcolsep}{5pt}
\renewcommand{\arraystretch}{1.1} %
\resizebox{\linewidth}{!}{
\begin{tabular}{l|c|c|c|c|c}
\Xhline{3\arrayrulewidth}
\multirow{1}{*}{} & \multirow{2}{*}{\textbf{Attn.}} & \multicolumn{4}{c}{\textbf{Prefill and Decode Sequence Lengths}} \\
\cline{3-6}
 & & \textbf{(340, 160)} & \textbf{(60, 20)} & \textbf{(7000, 200)} & \textbf{(1700, 400)} \\
\Xhline{3\arrayrulewidth}
\multirow{2}{*}{\textbf{\tabincell{c}{Latency\\(ms)}}} & Original & 325.00 & 40.61 & 709.59 & 894.21 \\ 
 & \textbf{Ours LA} & \textbf{290.08} & \textbf{37.51} & \textbf{432.48} & \textbf{736.51} \\ 
\Xhline{3\arrayrulewidth}
\multirow{2}{*}{\textbf{\tabincell{c}{Memory\\(GB)}}} & Original & 13.4 & 12.8 & 32.3 & 15.7 \\ 
 & \textbf{Ours LA} & \textbf{13.1} & \textbf{12.8} & \textbf{21.7} & \textbf{14.8} \\ 
\Xhline{3\arrayrulewidth}
\end{tabular}
}
\end{center}
\end{table}

We also provide detailed reports on latency and memory consumption for the LLaMA-2-7B model across four downstream tasks, under varied prefill and decode size configurations (see Appendix \ref{sec:more_profiling} for details). 
As shown in Tab.~\ref{tab:overall_comp_2}, our approach reduces latency by up to 39.1\% and memory usage by up to 32.8\% during runtime when deploying LLaMA-2-7B models on a A100-80G GPU.
\begin{table}[t]
\setlength\tabcolsep{4pt}
\caption{Accuracy comparison on six zero/few-shot downstream tasks under 0.8s latency (sequence length is 4K).}
\label{tab:overall_comp_3}
\begin{center}
\renewcommand{\arraystretch}{1.1} %
\resizebox{\linewidth}{!}{
\begin{tabular}{c|c|cccccc}
\Xhline{3\arrayrulewidth}
\textbf{LLaMA-2} & \textbf{Attn.} & \textbf{BBH} & \textbf{PIQA} & \textbf{MMLU} & \textbf{COPA} & \textbf{ARCC} & \textbf{AGNews} \\
\Xhline{3\arrayrulewidth}
7B & Original & 33.50 & 63.22 & \textbf{45.40} & 85.00 & \textbf{52.17} & 78.17 \\ \hline
13B & \textbf{Ours LA} & \textbf{33.91} & \textbf{68.06} & 36.57 & \textbf{85.00} & 51.74 & \textbf{78.95} \\
\Xhline{3\arrayrulewidth}
\end{tabular}
}
\end{center}
\end{table}

In addition, we compare the accuracy of our augmented linear attention method and the original attention-based LLaMA models under comparable inference latency on six downstream tasks.
As shown in Tab.~\ref{tab:overall_comp_3}, LLaMA-2-13B with our augmented linear attention, achieves comparable inference latency to the original LLaMA-2-7B at a 4K sequence length, while outperforming the original LLaMA-2-7B in four out of six downstream tasks. These results validate that our method can boost downstream task performance.

\subsection{Our Augmented LAs over SOTA LA Baselines}\label{sec:exp-main2}

\textbf{Overall Comparison.}
We apply our augmented LAs to five decoder-based or encoder-decoder-based LLMs and compare them with other LA baselines.
The training trajectories are visualized in Fig. \ref{fig:training_loss}. We see that our augmented LAs consistently achieve a better convergence loss as compared to all baselines. As for the quantitative results:
\begin{enumerate}[leftmargin=0.5cm]
    \item \textit{Text Classification with GPT-2 and T5.} We evaluate the performance of GPT-2 and T5 with our augmented LAs on the GLUE benchmark, with results provided in Appendix~\ref{appendix:overall_comp}. Our augmented LAs consistently yield better accuracy, achieving an average increase of 1.87 percentage points in classification accuracy on the GLUE benchmark compared to competitive existing LA baselines, such as FLASH-Local and FLASH-Global.
    \item \textit{Language Modeling with FLASH and LLaMA-7B/13B.} 
    We evaluate the perplexity of LLaMA-7B/13B with our augmented LAs on PG-19, with results provided in Appendix~\ref{appendix:overall_comp_2}. 
    The results on LLaMA models reveal that our augmented LAs with both the local augmentation and grouped LAs outperform all baselines, resulting in a 6.67/6.33 reduction in perplexity. 
    The results on FLASH models consistently validate the effectiveness of our augmented LAs, leading to 1.49 to 20.09 perplexity reductions as compared to other LAs and even 0.24 reduction over original attention.
    The effectiveness of our augmented LAs is consistently validated by results on FLASH models and the Wiki40B dataset, demonstrating perplexity reductions ranging from 1.49 to 20.09 as compared to baselines, and even a 0.24 reduction over the original attention.
    \item \textit{Downstream Tasks on LLaMA-2-7B.}
    We analysis six downstream tasks: BBH, PIQA, MMLU, COPA, ARCC, and AGNews. 
    Using standard evaluation settings, MMLU was tested with 5 shots, BBH with 3 shots, and the remaining tasks with zero shots. 
    As shown in Tab.~\ref{tab:downstream_task}, our augmented linear attention not only reduces perplexity but also improves accuracy across all tasks. Specifically, with models like FLASH, our method achieved an average accuracy improvement of 3.53\%.

\end{enumerate}
\begin{table}[t]
\setlength\tabcolsep{4pt}
\caption{Accuracy comparison using the LLaMA-2-7B model on six zero-shot or few-shot downstream tasks.}
\label{tab:downstream_task}
\begin{center}
\renewcommand{\arraystretch}{1.1} %
\resizebox{\linewidth}{!}{
\begin{tabular}{l|cccccc|c}
\Xhline{3\arrayrulewidth}
\textbf{Attn.} & \textbf{BBH} & \textbf{PIQA} & \textbf{MMLU} & \textbf{COPA} & \textbf{ARCC} & \textbf{AGNews} & \textbf{Lat.} \\
\Xhline{3\arrayrulewidth}
FLASH-Local & 30.89 & 61.65 & 34.21 & 68.60 & 45.01 & 69.05 & 0.5s \\ \hline
FLASH-Global & 31.78 & 61.48 & 35.62 & 75.00 & 47.36 & 78.14 & 0.5s \\ \hline
\textbf{Aug. FLASH} & \textbf{32.70} & \textbf{62.52} & \textbf{36.04} & \textbf{78.00} & \textbf{48.36} & \textbf{78.20} & 0.5s \\ 
\Xhline{3\arrayrulewidth}
\end{tabular}
}
\end{center}
\end{table}

In addition, we extend our methods to three more linear attention methods, with summarized results in Appendix \ref{appendix:more_la}.

\begin{table}[t]
\setlength\tabcolsep{4pt}
\caption{Throughput of LLaMA (tokens/s) with LAs and the speculative decoding on MT-Bench~\cite{zheng2023judging}.}
\label{tab:speculation}
\begin{center}
\renewcommand{\arraystretch}{1.1} %
\resizebox{\linewidth}{!}{
\begin{tabular}{l|cccc}
\Xhline{3\arrayrulewidth}
\textbf{LLaMA w/} & Loc. & Loc.+Gro. & Loc.+Conv & Loc.+Gro.+Conv  \\
\hline
7B & 32.3~\textbf{(1.0x)} & 26.8~\textbf{(1.0x)} & 30.4~\textbf{(1.0x)} & 25.9~\textbf{(1.0x)} \\ 
\textbf{7B w/ Spec.} & 63.3~\textbf{(2.0x)} & 50.5~\textbf{(1.9x)} & 55.1~\textbf{(1.8x)} & 50.7~\textbf{(2.0x)} \\
\hline
13B & 26.1~\textbf{(1.0x)} & 22.7~\textbf{(1.0x)} &  22.3~\textbf{(1.0x)} & 20.4~\textbf{(1.0x)} \\ 
\textbf{13B w/Spec.} & 54.4~\textbf{(2.1x)} & 42.6~\textbf{(1.9x)} & 47.0~\textbf{(2.1x)} & 41.7~\textbf{(2.0x)}  \\
\Xhline{3\arrayrulewidth}
\end{tabular}
}
\end{center}
\end{table}

\textbf{Generation Speedups by Integrating LAs with Speculative Decoding.}
We benchmark the speedups of our compatible LAs with speculative decoding.
As shown in Tab. \ref{tab:speculation}, we test the LLaMA-7B/13B models which are adapted into a chat model format, similar to LongLora~\cite{chen2023longlora}.
Following Medusa~\cite{cai2023medusa}, we train Medusa heads for speculative decoding. Speed tests for the 7B and 13B models are conducted on a single A100-80GB GPU, we observe that our revised LAs are compatible with speculative decoding and approximately doubled the speed.

\subsection{Ablation Study}
\label{sec:ablation}
\begin{table}[t]
\setlength\tabcolsep{4pt}
\caption{Comparison of our method with the integration of FLASH~\cite{flash} and Medusa~\cite{cai2023medusa}.}
\label{tab:direct_integration}
\begin{center}
\renewcommand{\arraystretch}{1.1} %
\resizebox{\linewidth}{!}{
\begin{tabular}{l|c|c|c|c}
\Xhline{3\arrayrulewidth}
\textbf{Methods} & \textbf{\tabincell{c}{Total Latency}} & \textbf{Attention} & \textbf{FFNs} & \textbf{Others} \\
\Xhline{3\arrayrulewidth}
FLASH + Medusa   & 137.2 ms & 119.7 ms & 8.2 ms & 9.3 ms \\ \hline
\textbf{Ours Aug. LA} & 49.7 ms \textbf{(-64\%)} & 32.2 ms & 8.2 ms & 9.3 ms \\ 
\Xhline{3\arrayrulewidth}
\end{tabular}
}
\end{center}
\end{table}

\textbf{Comparison with Direct Integration.}
To verify the effectiveness of our causal and compatible augmentation techniques, we compare them with the direct integration of previous linear attention FLASH~\cite{flash} and the speculative decoding method Medusa~\cite{cai2023medusa}.
As shown in Tab.~\ref{tab:direct_integration}, our method applied to LLaMA-2-7B models on A100 GPUs for a single batch of speculative decoding (64 speculated tokens and 42 sequence candidates), achieves a 64\% reduction in total latency compared to the direct integration, while also reducing QKV memory requirements by 75\% from 0.4 GB to 0.1 GB.

Our techniques outperform direct integration because standard implementations, even with linear attention like FLASH and speculative decoding like Medusa, face two key limitations without our augmentations: (1) slow sequence-based decoding and (2) lack of optimizations such as shared cumulative sum (\texttt{cumsum}) and key-value (KV) states for batch processing. 
Conventional strategies for compatible KV caching rely on sequence-based decoding, assigning distinct KV caches to each speculated sequence candidate, as shown in Fig.~\ref{fig:speculative_decoding}. This results in unnecessary computational effort and memory inefficiency since candidates with identical prefixes are processed separately.
In contrast, our method addresses these issues by ensuring identical prefixes are computed only once, mitigating these issues with time-dependent causal and compatible augmentation in linear attention and speculative decoding.

\textbf{Our LA Speedups.}
We benchmarked the training speed of FLASH using both the original attention and our augmented LAs, with a batch size of 1, on a single A100-40G GPU. 
Our results show that the augmented LAs significantly improve training speed. 
For sequence lengths of 4K and 8K, they are 1.52$\times$ and 2.94$\times$ faster, respectively.
FLASH with augmented LAs takes 1.05 seconds and 1.95 seconds per training step for 4K and 8K sequences, compared to 1.60 seconds and 5.74 seconds with the original attention.
The group size in FLASH was consistently set to 256.

\textbf{Extend to Longer Sequence.}
We fine-tuned LLaMA-2-7B to extend its sequence length from 4K to 8K using our augmented LAs, following LongLora~\cite{chen2023longlora} setting on the RedPajama dataset. For a fair comparison, we used only the local attention in LongLora, maintaining a block size of 256. Our augmented LAs reduced perplexity from 15.29 to 13.86, demonstrating their effectiveness in handling longer sequences.

\section{Conclusion}
\label{sec:conclusion}
This paper presents the first empirical analysis of linearized autoregressive LLMs, revealing significant limitations of existing linear attention methods in effectively handling masked attention and integration with speculative decoding.
To address these challenges, we introduced an approach that combines effective local augmentation with seamless compatibility for speculative decoding.
Our experiments across a range of LLMs consistently demonstrate that our method achieves substantial performance gains.
Notably, we achieve up to a 6.67 perplexity reduction and up to 2$\times$ speedups in generation compared to existing linear attention methods.
Our work paves the way for more efficient training and deployment of powerful autoregressive LLMs, especially for long-sequence applications.

\section*{Acknowledgements}
This work is supported by the National Science Foundation (NSF) EPCN program (Award number: 1934767) and the CoCoSys, one of the seven centers in JUMP 2.0, a Semiconductor Research Corporation (SRC) program sponsored by DARPA.
We extend our gratitude towards Arthur Szlam, Marc'aurelio Ranzato, and Cliff Young for reviewing the paper and providing insightful feedback. 
We also thank the extended team at Google DeepMind, who enabled and supported this research direction. 

\section*{Impact Statement}

\textbf{Efficient LLM Training and Serving Goal.}
The recent advancements in Large Language Models (LLMs), exemplified by OpenAI's GPT-3 with its 175 billion parameters, have underscored the significant data and computational power required for such technologies. 
Training models of this scale incur substantial costs, both financially and environmentally. 
For instance, the cost necessary to train GPT-3 could exceed 4 million equivalent GPU hours~\cite{gpt3}, and the carbon footprint of training a single Transformer model might rival the lifetime emissions of five average American cars~\cite{strubell2019energy}. 
Addressing the challenges of efficient training and serving of LLMs is therefore not only a technical imperative but also an environmental and ethical necessity.

\textbf{Societal Consequences.}
The success of this project in enabling more efficient training and serving of LLMs will have far-reaching implications, especially in processing long sequences commonly encountered in document handling. 
Our efforts are set to substantially influence various societal and economic sectors. The enhanced efficiency of LLMs promises transformative changes in diverse applications ranging from document summarization and question answering to personal digital assistants, security, and augmented reality. 
The development and exploration of linearized LLMs mark a pivotal progress in rendering these models both more accessible and environmentally sustainable.

\bibliography{ref}
\bibliographystyle{icml2024}

\newpage
\appendix
\onecolumn

\section{Comprehensive Related Works}
\label{appendix:related_work}

\textbf{Autoregressive LLMs.}
Transformers~\cite{transformer,vit} have significantly advanced the fields of language and vision, leading to the development of foundation LLMs such as ChatGPT~\cite{gpt3,gpt4}, LLaMA~\cite{touvron2023llama,touvron2023llama2}, Gemini~\cite{team2023gemini}, DALL-E~\cite{DALL-E}, etc.
To date, various Transformers have emerged to serve distinct needs, broadly categorized into three types: 
\textit{encoder-based}, \textit{decoder-based}, and \textit{encoder-decoder} models.
Encoder-based models like BERT~\cite{devlin2018bert} focus on natural language understanding and are also commonly used in image processing~\cite{vit}. 
Encoder-decoder models like the original Transformer~\cite{transformer}, Bard~\cite{bard}, and T5~\cite{2020t5,roberts2022t5x} are designed for sequence-to-sequence tasks (e.g., translation, speech recognition), where the encoder extracts features and the decoder produces outputs based on these features.
Decoder-based models, including GPT~\cite{gpt2,gpt4} and LLaMA~\cite{touvron2023llama}, generate text sequentially by predicting the next token based on previous ones. All these models leverage Transformer architectures but differ in their specific purposes and structures.
Both encoders and decoders are leveraged in multimodal models like MiniGPT~\cite{zhu2023minigpt,chen2023minigptv2} and DALL-E~\cite{DALL-E}.
Note that the model architectures used in all categories are based on Transformer. The primary difference lies in their purpose: the encoder is designed to extract features, while the decoder focuses on scoring and generating outputs.
Our work presents a comprehensive study of applying linear attention techniques to the encoder/decoder-based LLMs. 

\textbf{Efficient Linear Attention.}
Transformers' self-attention modules, known for their quadratic computational complexity~\cite{zhu2021long,katharopoulos2020transformers}, have spurred the development of linear attention methods to improve efficiency, especially in encoder-based LLMs for better training and inference. 
Techniques such as local attentions~\cite{liu2021swin,arar2022learned,wang2020linformer,tu2022maxvit,you2023vitcod} limit self-attention to neighboring tokens or group attention queries to reduce the computational cost, while kernel-based linear attentions~\cite{liu2021swin,arar2022learned,wang2020linformer,tu2022maxvit,you2024shiftaddvit} decompose the softmax with kernel functions and exchange the computation order. 
However, only a few linear attention approaches focus on decoder-based autoregressive LLMs, aiming to reduce RNN-style sequential state updates over a large number of steps~\cite {flash,katharopoulos2020transformers}.
Recent studies, like LongLoRA~\cite{chen2023longlora}, aim to adapt local attention techniques for efficient fine-tuning of pre-trained autoregressive LLMs, yet a thorough analysis comparing various linear attention methods for autoregressive LLMs remains lacking.
This paper uniquely provides a systematic review of existing linear attentions for decoder-based autoregressive LLMs and investigates how to efficiently enhance less effective linear attention methods.

\textbf{Speculative Decoding.}
Linear attention techniques alleviate the training inefficiency in LLMs by mitigating the quadratic complexity with regard to the number of input tokens. However, during deployment, autoregressive decoding necessitates sequential token-by-token text generation, which curtails parallelism and restricts the number of input tokens.
Speculative decoding~\cite{chen2023accelerating,miao2023specinfer,kim2023big,leviathan2023fast,cai2023medusa} has proven to be an effective strategy for boosting parallelism in LLM serving, utilizing small speculative models for initial generation, with original LLMs serving as validators to assess if the output meets standards or needs resampling. 
Recent works like Medusa~\cite{cai2023medusa} further argue that the small speculative models and LLMs can be the same model, and other studies~\cite{schuster2022confident,bae2023fast} suggest using shallow layers for generation and deeper layers for verification, based on early exit strategies.
Such speculative decoding and linear attention jointly ensure efficient LLM training and generation, especially for long sequence inputs.
In this paper, we take the initiative to investigate the synergy between linearized LLMs and speculative sampling, to improve the efficiency of training and serving LLMs. 

\section{More Visualization of Training Trajectories.}

\begin{figure}[ht]
    \centering
    \includegraphics[width=0.5\linewidth]{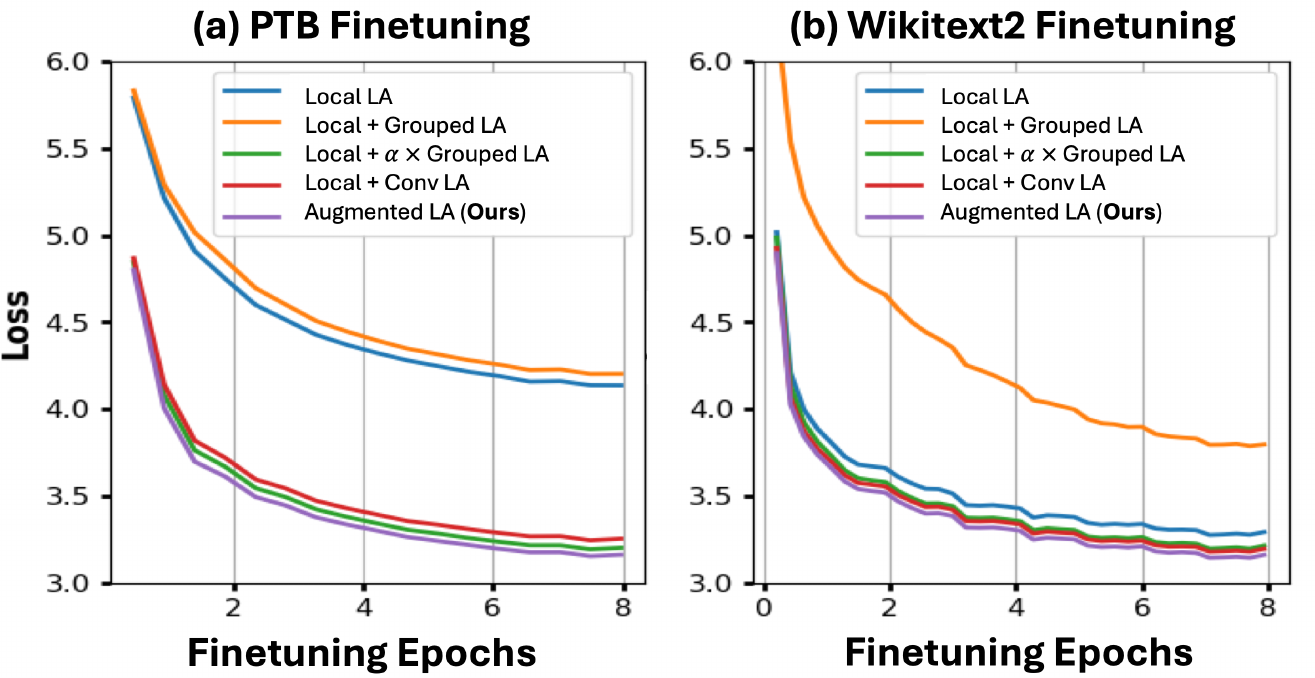}
    \caption{Visualizing the training trajectories of baseline LAs and our augmented LAs.}
    \label{fig:gpt-2-loss}
\end{figure}

As detailed in Sec. \ref{sec:ablation}, we present a quantitative analysis comparing local LAs, grouped LAs, and our augmented LAs that combine both local augmentation and grouped LAs. This appendix provides the training trajectories for GPT-2 using these LA methods. Fig. \ref{fig:gpt-2-loss} demonstrates that our local augmentation, specifically masked DWConv, effectively enhances both local and grouped LAs. Moreover, our augmented LAs, which integrate local augmentation with grouped LAs, exhibit the most favorable convergence in terms of loss.

\section{More Profiling on the LLaMA-2-7B Model}
\label{sec:more_profiling}

We provide detailed profiling and comparisons below to illustrate the runtime distribution between attention and feed-forward networks (FFNs), highlighting that attention is a bottleneck even for LLMs with 7B parameters.
To ensure a real-world application scenario, we profiled the LLaMA-2-7B model across four settings of prefill and decode sizes, adhering to benchmarks commonly used in academia and industry, as summarized in Tab.~\ref{tab:profiling_tasks}.

\begin{table}[ht]
\setlength\tabcolsep{4pt}
\caption{Dataset and task details for different prefill and decode size settings.}
\label{tab:profiling_tasks}
\begin{center}
\renewcommand{\arraystretch}{1.1} %
\resizebox{0.6\linewidth}{!}{
\begin{tabular}{l|l|l|c}
\Xhline{3\arrayrulewidth}
\textbf{(Prefill, Decode)} & \textbf{Task} & \textbf{Dataset} & \textbf{Referenced Paper} \\
\Xhline{3\arrayrulewidth}
(340, 160) & Chat & ShareGPT & \cite{kwon2023efficient} \\ \hline
(60, 20) & Chat & Stanford Alpaca & \cite{taori2023stanford} \\ \hline
(7000, 200) & Summarization & ArXiv Summarization & \cite{agrawal2024taming} \\ \hline
(1700, 400) & Chat & OpenChat ShareGPT 4 & \cite{agrawal2024taming} \\ 
\Xhline{3\arrayrulewidth}
\end{tabular}
}
\end{center}
\end{table}

As shown in Tab.~\ref{tab:profiling_results}, profiling the LLaMA-2-7B models under the four prefill and decode size settings reveals that the average runtime latency attributed to attention and FFNs accounts for 55\% and 21\% of the total runtime across these settings, respectively. This indicates that although FFNs are a bottleneck in the model, attention is an even more significant bottleneck, especially for large-scale LLMs and extended dialogue sequences (e.g., 67.8\% runtime latency for the arxiv summarization task). Therefore, optimizing attention blocks can yield considerable speed improvements, particularly for tasks with large prefill or decode sequence lengths. This is corroborated by contemporary studies on linear attention-based LLMs~\cite{lee2023sea,yang2023gated} and efforts to optimize attention, such as FlashAttention~\cite{dao2022flashattention} and FLAT~\cite{flat}.

\begin{table}[ht]
\setlength\tabcolsep{4pt}
\caption{Latency breakdown of LLaMA-2-7B models under different prefill and decode size settings.}
\label{tab:profiling_results}
\begin{center}
\renewcommand{\arraystretch}{1.1} %
\resizebox{0.7\linewidth}{!}{
\begin{tabular}{l|c|c|c|c}
\Xhline{3\arrayrulewidth}
\textbf{(Prefill, Decode)} & \textbf{(340, 160)} & \textbf{(60, 20)} & \textbf{(7000, 200)} & \textbf{(1700, 400)} \\
\Xhline{3\arrayrulewidth}
Attention (ms) & 158.97 \textbf{(48.9\%)} & 20.12 \textbf{(49.5\%)} & 481.35 \textbf{(67.8\%)} & 481.41 \textbf{(53.8\%)} \\ 
FFNs (ms) & 74.64 \textbf{(23.0\%)} & 9.22 \textbf{(22.7\%)} & 111.90 \textbf{(15.8\%)} & 188.98 \textbf{(21.1\%)} \\ 
Others (ms) & 91.39 \textbf{(28.1\%)} & 11.27 \textbf{(27.8\%)} & 116.34 \textbf{(16.4\%)} & 223.83 \textbf{(25.1\%)} \\ 
\hline
Total Latency (ms) & 325.00 & 40.61 & 709.59 & 894.21 \\ 
\Xhline{3\arrayrulewidth}
\end{tabular}
}
\end{center}
\end{table}

\section{Breakdown Analysis of Our Augmented LAs}

To gain insights into the contribution of each component in our augmented LAs, we show the breakdown analysis using GPT-2 and T5 models on Wikitext2~\cite{merity2017pointer}/PTB~\cite{marcus1993building} and CNN/Daily Mail~\cite{see2017get} datasets, respectively. 
As shown in Tabs.~\ref{tab:gpt2_ablation} and \ref{tab:t5_ablation}, 
our local augmentation, i.e., masked DWConv, consistently augments the local or grouped LAs, leading to 5.71 perplexity reductions on GPT-2 and 3.59 Rouge1 score~\cite{lin-2004-rouge} improvement on T5.
Our augmented LAs, consisting of both local augmentation and grouped LAs, achieve the best results, i.e., 11.83$\sim$17.54 perplexity reduction and 4.23$\sim$15.45 Rouge1 score improvement, over all other LA variants.

\begin{table}[ht]
\setlength\tabcolsep{4pt}
\caption{Perplexity of GPT-2 with our augmented LAs on the Wikitext2 and PTB datasets.}
\label{tab:gpt2_ablation}
\begin{center}
\renewcommand{\arraystretch}{1.1} %
\resizebox{0.56\linewidth}{!}{
\begin{tabular}{l|ccc|c}
\Xhline{3\arrayrulewidth}
\textbf{GPT-2 w/} & Loc. & Loc.+Gro. & Loc.\textbf{+Conv} & \textbf{Augmented LA}  \\
\Xhline{3\arrayrulewidth}
Wikitext2 & 56.80  &  42.81 &  51.09  & 39.26 \\ 
PTB & 69.32 &  57.72 & 84.24 & 46.85 \\
\Xhline{3\arrayrulewidth}
\end{tabular}
}
\end{center}
\end{table}

\begin{table}[ht]
\setlength\tabcolsep{2pt}
\caption{Ablation studies of fine-tuning T5 with LAs on the CNN/Daily Mail dataset~\cite{see2017get}.}
\label{tab:t5_ablation}
\begin{center}
\renewcommand{\arraystretch}{1.1} %
\resizebox{0.56\linewidth}{!}{
\begin{tabular}{p{3.5cm}|p{1.8cm}|p{1.8cm}|p{1.8cm}|p{1.8cm}}
\Xhline{3\arrayrulewidth}
\textbf{T5 w/} & \multicolumn{1}{c|}{\textbf{Rouge1}} & \multicolumn{1}{c|}{\textbf{Rouge2}} & \multicolumn{1}{c|}{\textbf{RougeL}} & \multicolumn{1}{c}{\textbf{RougeLsum}}  \\
\Xhline{3\arrayrulewidth}
Local LA& \multicolumn{1}{c|}{8.65} & \multicolumn{1}{c|}{0.17} & \multicolumn{1}{c|}{7.14} & \multicolumn{1}{c}{8.27} \\ 
Grouped LA& \multicolumn{1}{c|}{6.14}  & \multicolumn{1}{c|}{0.86} & \multicolumn{1}{c|}{5.77} & \multicolumn{1}{c}{5.50} \\
Local + Grouped LA& \multicolumn{1}{c|}{19.87} & \multicolumn{1}{c|}{3.07} & \multicolumn{1}{c|}{14.54} &  \multicolumn{1}{c}{18.29} \\
Local + $\alpha \times$Grouped LA& \multicolumn{1}{c|}{19.01} & \multicolumn{1}{c|}{2.90} & \multicolumn{1}{c|}{13.99} & \multicolumn{1}{c}{17.54} \\ 
Local LA + DWConv& \multicolumn{1}{c|}{12.24} & \multicolumn{1}{c|}{0.20} & \multicolumn{1}{c|}{8.95} & \multicolumn{1}{c}{11.38} \\
\hline
\textbf{Augmented LAs} & \multicolumn{1}{c|}{24.10} & \multicolumn{1}{c|}{4.93} & \multicolumn{1}{c|}{17.22} & \multicolumn{1}{c}{22.11} \\ 
\Xhline{3\arrayrulewidth}
\end{tabular}
}
\end{center}
\end{table}

\section{Additional Training and Evaluation Settings and Model Hyperparameters.}

\textbf{Model, Task, Dataset.} \textit{Model:} We evaluate seven existing linear attention methods on top of three representative Transformers: (1) encoder-based BERT model of 12 layers and around 400M parameters, (2) decoder-based GPT-2 model of 12 layers and around 500M parameters, and (3) encoder-decoder-based T5 model of 12 layers and around 900M parameters. \textit{Task and Dataset:} We conduct the evaluation on the text classification task across seven linguistic tasks from the General Language Understanding Evaluation (GLUE) benchmark: SST2, WNLI, QNLI, MNLI, RTE, MRPC, and QQP.

\textbf{Training and Evaluation Settings.} We fine-tuned all models for 3 epochs with a sequence length of 256, using a learning rate of $2 \times 10^{-5}$ and a batch size of 32. The optimizer was AdamW, with $\beta_1 = 0.9$ and $\beta_2 = 0.95$ following the standard training recipe in ~\cite{devlin2018bert}. For the GLUE task, classification accuracy was utilized to evaluate the performance of all linear attention methods.

\section{Text Classification with GPT-2 and T5}
\label{appendix:overall_comp}

We evaluate the performance of GPT-2 and T5 with our augmented LAs on the GLUE benchmark. As shown in Tab.~\ref{tab:gput_t5_results}, our augmented LAs consistently yield better accuracy, achieving an average increase of 1.87 percentage points in classification accuracy on the GLUE benchmark compared to competitive existing LA baselines, such as FLASH-Local/Global.

\begin{table}[ht]
\setlength\tabcolsep{3pt}
\caption{
Evaluation of augmented LAs on T5 and GPT-2, with the classification accuracy on the GLUE benchmark.
}
\begin{center}
\renewcommand{\arraystretch}{1.1} %
\resizebox{0.7\linewidth}{!}{
\begin{tabular}{l|ccccccc|c}
\Xhline{3\arrayrulewidth}
\textbf{GPT-2 w/ } & \textbf{SST2} & \textbf{WNLI} & \textbf{QNLI} & \textbf{MNLI} & \textbf{RTE} & \textbf{MRPC} & \textbf{QQP} & \textbf{Average}\\
\hline
\textbf{LA Baseline}    & 83.60&	53.52	& 77.16	& 73.97	&48.01	& 68.87	& 86.40&	\textbf{70.22}   \\
\hline
\textbf{Loc.+Gro.}    &  82.34	& 46.48	& 79.11 & 75.09 &	50.20	& 68.38	& 86.16	& 69.68   \\
\textbf{Loc.+$\alpha$*Gro.}    &  83.72	& 54.04	& 79.15 & 73.76 &	46.68	& 69.61	& 86.11	& 70.44   \\
\textbf{Augmented LA} & 84.72 	& 54.93	& 80.01		& 74.26 	& 50.90	& 69.85	& 86.16 & \textbf{71.55} \\

\hline
\hline
\textbf{T5 w/ } & \textbf{SST2} & \textbf{WNLI} & \textbf{QNLI} & \textbf{MNLI} & \textbf{RTE} & \textbf{MRPC} & \textbf{QQP} & \textbf{Average}\\
\hline
\textbf{LA Baseline}    &  77.87	& 56.34	& 58.87 & 49.44&	52.71	& 68.38	&75.62	& \textbf{62.75}   \\
\hline
\textbf{Loc.+Gro.}    &   76.95	& 56.34	& 60.37 & 51.20 &	49.82	& 68.38	& 75.44	&  62.64   \\
\textbf{Loc.+$\alpha$*Gro.}    &  78.10	& 56.34	& 59.62 & 51.49 &	49.10	& 68.38	& 75.62	& 62.66   \\
\textbf{Augmented LA} & 82.00 	& 56.34	& 59.78		&  54.26 	& 54.15	& 68.38	& 76.68 & \textbf{64.51} \\

\hline

\end{tabular}
}
\end{center}
\label{tab:gput_t5_results}
\end{table}

\section{Language Modeling with FLASH and LLaMA-7B/13B} 
\label{appendix:overall_comp_2}

We evaluate the perplexity of LLaMA-7B/13B with our augmented LAs on PG-19. As shown in Tab.~\ref{tab:ppl}, integrating our local augmentation, i.e., masked DWConv, with the local LAs results in a 6.67/6.33 reduction in perplexity. 
The results on LLaMA models reveal that our augmented LAs with both the local augmentation and grouped LAs outperform all baselines, resulting in a 6.67/6.33 reduction in perplexity. 
The results on FLASH models and the Wiki40B dataset consistently validate the effectiveness of our augmented LAs, leading to 1.49 to 20.09 perplexity reductions as compared to other LAs and even 0.24 reduction over original attention.
The effectiveness of our augmented LAs is consistently validated by results on FLASH models and the Wiki40B dataset, demonstrating perplexity reductions ranging from 1.49 to 20.09 as compared to baselines, and even a 0.24 reduction over the original attention.

\begin{table}[ht]
\setlength\tabcolsep{4pt}
\caption{Perplexity evaluation on two tasks: (1) LLaMA models on PG-19 (sequence length is 4K) and (2) FLASH model on Wiki40B (sequence length is 1K).}
\label{tab:ppl}
\begin{center}
\renewcommand{\arraystretch}{1.1} %
\resizebox{0.6\linewidth}{!}{
\begin{tabular}{l|p{0.5cm}|p{1.1cm}|p{0.5cm}|p{0.6cm}|p{1.7cm}}
\Xhline{3\arrayrulewidth}
Model & \multicolumn{1}{c|}{Loc.} & \multicolumn{1}{c|}{Loc.+Gro.} & \multicolumn{2}{c|}{Loc.\textbf{+Conv}} & \multicolumn{1}{c}{\textbf{Augmented LA}}   \\
\hline
LLaMA-2-7B & \multicolumn{1}{c|}{21.61}  &  \multicolumn{1}{c|}{15.04} &  \multicolumn{2}{c|}{14.94}  & \multicolumn{1}{c}{\textbf{13.47}} \\ 
LLaMA-2-13B & \multicolumn{1}{c|}{19.25} &  \multicolumn{1}{c|}{12.92} & \multicolumn{2}{c|}{12.92} & \multicolumn{1}{c}{\textbf{11.55}} \\
\hline
\hline
Model & \multicolumn{1}{c|}{Loc.} & \multicolumn{1}{c|}{Loc.+Gro.}  & \multicolumn{1}{c|}{Gro.} & \multicolumn{1}{c|}{Quad.} & \multicolumn{1}{c}{\textbf{Augmented LA}}  \\
\hline
FLASH-110M & \multicolumn{1}{c|}{16.65} & \multicolumn{1}{c|}{16.14} & \multicolumn{1}{c|}{35.25} & \multicolumn{1}{c|}{15.40} & \multicolumn{1}{c}{\textbf{15.16}}  \\
\Xhline{3\arrayrulewidth}
\end{tabular}
}
\end{center}
\end{table}

\section{Augmentation for More Linear Attention Methods}
\label{appendix:more_la}
We further extend our augmentation method to four additional types of linear attention methods, including not only FLASH but also the random feature attention (RFA)~\cite{peng2021random}, Performer~\cite{performers}, and Linformer~\cite{wang2020linformer}. Specifically, we evaluated these linear attention methods before and after our augmentation on the decoder-based GPT-2 model and measured the resulting text classification accuracy on the GLUE benchmark~\cite{wang2018glue}. Tab.~\ref{tab:more_LAs} demonstrates that our augmentation method consistently improves performance across these methods, achieving non-trivially on average 5.07\% $\sim$ 8.05\% task accuracy gain. These results validate that our augmentation techniques are generally applicable to different linear attention methods in largely enhancing their achievable performance and efficiency.

\begin{table}[ht]
\setlength\tabcolsep{4pt}
\caption{Augmentation for various linear attention methods.}
\label{tab:more_LAs}
\begin{center}
\renewcommand{\arraystretch}{1.1} %
\resizebox{0.7\linewidth}{!}{
\begin{tabular}{l|ccccccc|c}
\Xhline{3\arrayrulewidth}
\textbf{Method} & \textbf{SST2} & \textbf{RTE} & \textbf{MRPC} & \textbf{QQP} & \textbf{MNLI} & \textbf{QNLI} & \textbf{WNLI} & \textbf{Average} \\
\Xhline{3\arrayrulewidth}
RFA & 83.37 & 61.01 & 73.04 & 82.24 & 71.73 & 69.45 & 45.07 & 69.42 \\ 
\textbf{Aug. RFA} & 91.28 & 60.65 & 75.00 & 88.53 & 81.76 & 69.26 & 54.93 & \textbf{74.49} \\ \hline
Performer & 86.93 & 49.46 & 69.12 & 76.30 & 70.60 & 69.36 & 38.03 & 65.69 \\ 
\textbf{Aug. Performer} & 91.51 & 67.15 & 72.30 & 84.61 & 70.87 & 63.72 & 50.70 & \textbf{71.55} \\ \hline
Linformer & 79.47 & 52.35 & 68.38 & 76.30 & 34.56 & 69.06 & 52.11 & 60.59 \\ 
\textbf{Aug. Linformer} & 92.43 & 61.37 & 77.45 & 88.42 & 42.63 & 63.26 & 54.93 & \textbf{68.64} \\ 
\Xhline{3\arrayrulewidth}
\end{tabular}
}
\end{center}
\end{table}

\end{document}